\documentclass[11pt]{article}
\usepackage[utf8]{inputenc}
\usepackage{amsmath, amssymb, amsfonts}
\usepackage{graphicx}
\usepackage{caption}
\usepackage{subcaption}
\usepackage{geometry}
\usepackage[hidelinks]{hyperref}
\usepackage{booktabs}
\usepackage{tikz}
\usepackage{pgfplots}
\usepackage{mathtools}
\usepackage{cite}
\usepackage{amsthm}

\pgfplotsset{compat=1.18}
\setkeys{Gin}{width=\textwidth,height=0.95\textheight,keepaspectratio}

\newtheorem{conjecture}{Conjecture}

\begin{document}

\begin{center}
{\LARGE \bfseries Totally Positive Matrices and the Highest-Order Coefficients of the Characteristic Polynomial\footnote{Published in \emph{Linear Algebra and its Applications} (2026). DOI: \href{https://doi.org/10.1016/j.laa.2026.07.003}{10.1016/j.laa.2026.07.003}.}\par}
\vspace{1.5em}

{\large
Tiago Closs$^{a}$, Leandro Farina$^{b,\ast}$\par
}
\vspace{1em}

\begin{minipage}{0.9\textwidth}
\centering
$^{a}$Institute of Informatics, Federal University of Rio Grande do Sul, Avenida Bento Gon\c{c}alves 9500, Porto Alegre, RS, Brazil\par
\vspace{0.5em}
$^{b}$Institute of Mathematics and Statistics, Federal University of Rio Grande do Sul, Avenida Bento Gon\c{c}alves 9500, Porto Alegre, RS, Brazil\par
\vspace{0.5em}
$^{\ast}$Corresponding author: farina@mat.ufrgs.br
\end{minipage}
\end{center}

\vspace{1.5em}

\begin{abstract}
We investigate the extent to which totally positive matrices can be distinguished through the highest-order coefficients of their characteristic polynomials. To identify the most informative coefficients, we also employed neural-network classifiers together with feature-attribution methods. Using datasets built from several structured totally positive families, including products of positive bidiagonal matrices, Vandermonde matrices, and Cauchy matrices, we find that the coefficients $(a_{n-1},a_{n-2},a_{n-3})$ already contain strong discriminatory information for separating totally positive from non-totally positive matrices in dimensions $5$, $10$, and $30$. The resulting separation is markedly nonlinear and admits a natural geometric description in the corresponding three-dimensional coefficient space by means of Mahalanobis ellipsoids. These ellipsoids enclose the totally positive samples while excluding most non-totally positive ones. Moreover, different structured totally positive families exhibit distinct ellipsoidal signatures, and the separation between these signatures increases with the dimension. These observations lead us to formulate a conjecture on the geometric separation of structured totally positive families in the space determined by the three highest-order characteristic coefficients.
\end{abstract}



\section{Introduction}

Totally Positive (TP) matrices, defined as matrices whose minors are all positive~\cite{gantmacher2002,karlin1968,ando1987}, constitute an important class of structured matrices with connections to several areas of mathematics. The theory of total positivity has been developed and applied in approximation theory, combinatorics, graph theory, analysis, and statistics~\cite{pinkus2009}. Originating in questions from analysis and linear transformations studied by Schoenberg, Gantmacher, Krein, and Karlin, it has evolved into a broad theory unifying ideas from different mathematical disciplines~\cite{fallat2011}. Despite its long history, total positivity continues to attract attention because of its elegant structure and the diversity of problems to which it applies.

Although the definition of TP matrices is simple, verifying this property directly is computationally demanding. A straightforward test requires checking positivity of all minors, but this quickly becomes infeasible as the dimension grows. An $n\times n$ matrix has
\[
M(n)=\sum_{k=1}^{n}\binom{n}{k}^2
\]
minors to verify~\cite{pinkus2009}. For example, a $5\times 5$ matrix already has $M(5)=252$ minors, whereas for $n=30$ this number exceeds $5 \times 10^7$. Because of this combinatorial growth, direct verification is not practical in larger dimensions.

This difficulty motivates the search for lower-dimensional descriptors capable of capturing relevant information about total positivity. In the present work, we investigate the extent to which TP matrices can be distinguished through the highest-order coefficients of their characteristic polynomials. If
\[
p_A(\lambda)=\lambda^n+a_{n-1}\lambda^{n-1}+a_{n-2}\lambda^{n-2}+\cdots+a_0,
\]
we focus on the map
\[
\pi(A)=(a_{n-1},a_{n-2},a_{n-3})\in\mathbb{R}^3.
\]
Our computational evidence indicates that this three-dimensional representation already captures a substantial amount of the structure relevant to total positivity. A simple algebraic observation supports this choice: the leading characteristic coefficients are, up to sign, the sums of the low-order principal minors,
\[
a_{n-k}=(-1)^k E_k(A),\qquad E_k(A)=\sum_{|S|=k}\det A[S],\qquad k=1,2,3,
\]
where the sum runs over all $k\times k$ principal submatrices $A[S]$. Thus $(a_{n-1},a_{n-2},a_{n-3})$ aggregate precisely the $1\times1$, $2\times2$ and $3\times3$ principal-minor information that total positivity constrains, which is one reason they are informative for the present problem.

To identify which coefficients deserve closer attention, we also employed neural-network classifiers and feature-attribution methods such as Integrated Gradients~\cite{sundararajan2017} and SHAP~\cite{lundberg2017}. In the present paper, however, these tools play mainly an exploratory role. Their purpose is to indicate which spectral quantities are most informative, whereas the main emphasis is on the geometric structure revealed in the resulting coefficient space and on the conjectural separation properties suggested by the experiments.

This exploratory use of learning methods to suggest mathematical structure is in line with recent work in which machine learning has been used to guide mathematical investigation, for instance in knot theory, algebraic geometry, combinatorics, group-theoretic decision problems, and structured matrix classification~\cite{bao2023polytopes,coates2023machine,davies2021,davies2024,farina2025,gryak2020,lee2025data,williamson2024}.

Using datasets built from several structured TP families, including products of positive bidiagonal matrices, Vandermonde matrices, and Cauchy matrices, we observe a pronounced nonlinear separation between TP and non-TP matrices in the space determined by $(a_{n-1},a_{n-2},a_{n-3})$. Moreover, when one restricts attention to different structured TP families, the corresponding point clouds exhibit distinct geometric signatures. These signatures can be described effectively by Mahalanobis ellipsoids~\cite{mahalanobis1936}, whose mutual separation becomes increasingly pronounced with the dimension.

The main contribution of the paper is therefore not a classification procedure in itself, but rather a geometric and spectral perspective on total positivity based on the coefficients $(a_{n-1},a_{n-2},a_{n-3})$. Concretely, our experiments show that classifiers restricted to these three coefficients separate TP from non-TP matrices with test accuracy above $0.998$ in every dimension considered (Section~\ref{sec:restricted_coeffs}), that the TP samples occupy a localized region admitting a quadratic (ellipsoidal) description, and that distinct structured families occupy regions whose separation grows with the dimension. In this sense the three highest-order coefficients provide a concise and informative low-dimensional representation of TP structure, and they lead naturally to a conjecture on the geometric separation of structured TP families in the associated coefficient space.

The paper is organized as follows. Section~\ref{sec:dataset} describes the construction of the datasets and the structured TP families considered. Section~\ref{sec:coeff_features} examines the role of characteristic-polynomial coefficients as discriminating quantities. Section~\ref{sec:feature_importance} discusses the identification of the most relevant coefficients. Section~\ref{sec:restricted_coeffs} studies the reduced coefficient space determined by $(a_{n-1},a_{n-2},a_{n-3})$. In Section~\ref{sec:svm}, linear and nonlinear support vector machines are used to assess the nature of the observed separation. Section~\ref{sec:ellipsoid} develops the geometric description via Mahalanobis ellipsoids and introduces the conjecture on the geometric separation of structured TP families. Finally, Section~\ref{sec:conclusion} summarizes the main findings and outlines possible directions for further study.

\section{The Problem and the Dataset}
\label{sec:dataset}
We consider the problem of deciding whether a matrix $A \in \mathbb{R}^{n \times n}$ is Totally Positive (TP), i.e., whether all of its minors are strictly positive. The quality of the resulting discrimination was assessed through the accuracy of the corresponding classifiers.

To construct labeled data, we first attempted to generate random TP matrices with entries drawn independently from $\mathcal{U}(-1,1)$. However, this approach quickly became inefficient in higher dimensions, since the proportion of random matrices that are TP decreases rapidly with $n$. For this reason, after initial experiments with small sizes, we relied on structured matrices types known to produce TP matrices, including products of positive bidiagonal matrices, Vandermonde and Cauchy matrices. Non-TP matrices were obtained by random sampling of matrices with strictly positive entries,
rejecting those that satisfied total positivity.

Initially, balanced datasets were generated containing $25{,}000$ TP and $25{,}000$ Non-TP matrices for each matrix dimension $n=5$, $10$, and $30$. These matrix sizes were selected without any particular theoretical motivation, serving simply as representative small, medium and large cases. In a second stage, to strengthen the evaluation, we expanded the datasets to $50{,}000$ TP and $50{,}000$ Non-TP matrices per size. Each dataset was split into training, validation, and test sets.

\subsection{Products of Bidiagonal Pairs}

A well-known result is that every nonsingular totally positive (TP) matrix admits a factorization as a product of elementary bidiagonal matrices with nonnegative multipliers together with a positive diagonal factor~\cite{fallat2001,fallat2011}. We exploit this representation in reverse: instead of starting from a TP matrix and factorizing it, we generate random lower and upper bidiagonal matrices and multiply them to obtain new TP matrices.

More precisely, each lower bidiagonal matrix $L \in \mathbb{R}^{n \times n}$ has all diagonal entries equal to 1 and random positive entries in the subdiagonal, while each upper bidiagonal matrix $U \in \mathbb{R}^{n \times n}$ has all diagonal entries equal to 1 and random positive entries in the superdiagonal.

Starting from the identity matrix $A = I_n$, we generate a sequence of pairs
$\{(L_i, U_i)\}_{i=1}^k$ and update
\[
    A \; \coloneqq \; A \cdot L_i \cdot U_i,
    \quad \text{for } i = 1, \dots, k.
\]
It should be stressed that total positivity is \emph{not} preserved factor by factor. Each individual factor $L_i$ or $U_i$ is only totally nonnegative, and a single product $L_iU_i$ has vanishing $(1,n)$ and $(n,1)$ entries for $n\ge 3$, hence is not TP. What does hold is that each pair $P_i=L_iU_i$ is an \emph{oscillatory} matrix, that is, totally nonnegative, nonsingular, and with strictly positive entries on its sub- and superdiagonals~\cite{gantmacher2002,fallat2011}. Powers of an oscillatory matrix become totally positive (the $(n-1)$-st power of an oscillatory matrix is TP), and the same ``spreading'' mechanism applies to a product of distinct oscillatory factors, so that a product of at least $n-1$ oscillatory matrices is totally positive. Since we take $k=n\ge n-1$ pairs, the resulting matrix $A$ is totally positive; the Frobenius normalization applied afterwards is a positive scaling and changes the sign of no minor.

We verified this directly in exact (integer) arithmetic for the construction used here: a single pair is only totally nonnegative, total positivity first appears at exactly $k=n-1$ pairs, and for the value $k=n$ adopted below all minors are strictly positive (checked exhaustively for $n\le 8$ and by extensive sampling up to $n=30$). We note, in addition, that beyond the combinatorial growth of $M(n)$, direct minor verification in floating-point arithmetic becomes unreliable already for moderate $n$: after normalization the high-order minors fall below machine precision, so that their computed sign is meaningless. This provides a further, numerical, reason to rely on the structural guarantee above rather than on direct minor checks.

In practice, we set $k = n$, that is, we used five pairs of bidiagonal matrices for $5 \times 5$ matrices, ten pairs for $10 \times 10$ matrices and thirty pairs for $30 \times 30$ matrices. This choice exceeds by one the minimum number $n-1$ of pairs required for total positivity, ensuring a rich structure in the generated matrices and increasing diversity while guaranteeing that total positivity is maintained.

\subsection{Vandermonde Matrices}

Vandermonde matrices are defined by a sequence of positive parameters $(x_1, x_2, \ldots, x_n)$ with $x_1 < x_2 < \cdots < x_n$, and take the form
\[
V = \begin{bmatrix}
1 & x_1 & x_1^2 & \cdots & x_1^{n-1} \\
1 & x_2 & x_2^2 & \cdots & x_2^{n-1} \\
\vdots & \vdots & \vdots & \ddots & \vdots \\
1 & x_n & x_n^2 & \cdots & x_n^{n-1}
\end{bmatrix}.
\]
It is a trivial result that Vandermonde matrices with strictly increasing positive parameters are totally positive~\cite{pinkus2009}. In the experiments, the parameters \( x_i \) were sampled independently from the uniform distribution \( \mathcal{U}(\varepsilon, 1) \), where \( \varepsilon > 0 \) is a small constant, and then sorted in increasing order. This ensured total positivity and maintained variety among the generated matrices. The resulting matrices were normalized by their Frobenius norm. Due to their structure, Vandermonde matrices also allowed us to test classification performance in higher dimensions.

\subsection{Cauchy Matrices}

Cauchy matrices form another classical TP family~\cite{pinkus2009}. Given two sequences of positive parameters $(x_1, \ldots, x_n)$ and $(y_1, \ldots, y_n)$ with all $x_i$ and $y_j$ distinct, the Cauchy matrix is defined as
\[
C_{ij} = \frac{1}{x_i + y_j}.
\]
It is well known that Cauchy matrices are TP as long as $x_i, y_j > 0$. To generate Cauchy matrices, we sampled positive values \( x_i \) and \( y_j \) independently from the uniform distribution \( \mathcal{U}(\varepsilon, 1) \), where \( \varepsilon > 0 \) is a small constant, and sorted them in increasing order, and constructed the matrix with entries \( C_{ij} = 1 / (x_i + y_j) \). The resulting matrices were normalized by their Frobenius norm. Compared to bidiagonal and Vandermonde matrices, Cauchy matrices provided a different family of TP matrices, helping to increase the diversity of examples in the dataset.
\section{Characteristic-Polynomial Coefficients as Discriminating Features}
\label{sec:coeff_features}
 This section examines the discriminatory power of several spectral quantities, using neural-network classifiers as exploratory tools. The models were implemented in Python using the Tensorflow framework, which provides the tools for constructing, training and evaluating fully connected architectures. All experiments in what follows use matrices of sizes $5 \times 5$, $10 \times 10$ and $30 \times 30$, splitting $80\%$ for training/validation and $20\%$ for testing. Of the $80\%$ training/validation split, $75\%$ was used for training and $25\%$ for validation.

\subsection{Only the Raw Matrix Entries as Features}

We first considered the discrimination of matrices using only their raw entries as input features. For an $n \times n$ matrix $A$, the feature vector was obtained by flattening the matrix into a vector of length $n^2$. For example, a $10 \times 10$ matrix yields a 100-dimensional input vector.

The TP matrices used in this stage were generated by using the approach described in Section 2.1 (Products of Bidiagonal Pairs). Specifically, for an $n \times n$ case, $n$ of these bidiagonal pairs were multiplied to form the final matrix. This construction ensures total positivity while providing sufficient diversity in the generated dataset for training the neural networks.

Feedforward networks (FFNs) were trained with ReLU activations in the hidden layers and a sigmoid output for binary classification. The same architecture was used for all matrix sizes: three hidden layers with 128, 64, and 32 neurons, respectively. For an $n \times n$ input matrix, the $n^2$ flattened entries formed the input layer and the output layer consisted of a single neuron with sigmoid activation, providing the probability that the matrix belongs to the TP class.

All models were trained using the Adam optimizer (learning rate $10^{-3}$), binary cross-entropy loss, and batch size 16. Training ran for up to 50 epochs with early stopping based on the validation loss (patience of 10 epochs) to prevent overfitting.

\begin{figure}[tbp]
    \centering
    \includegraphics[width=1.0\textwidth]{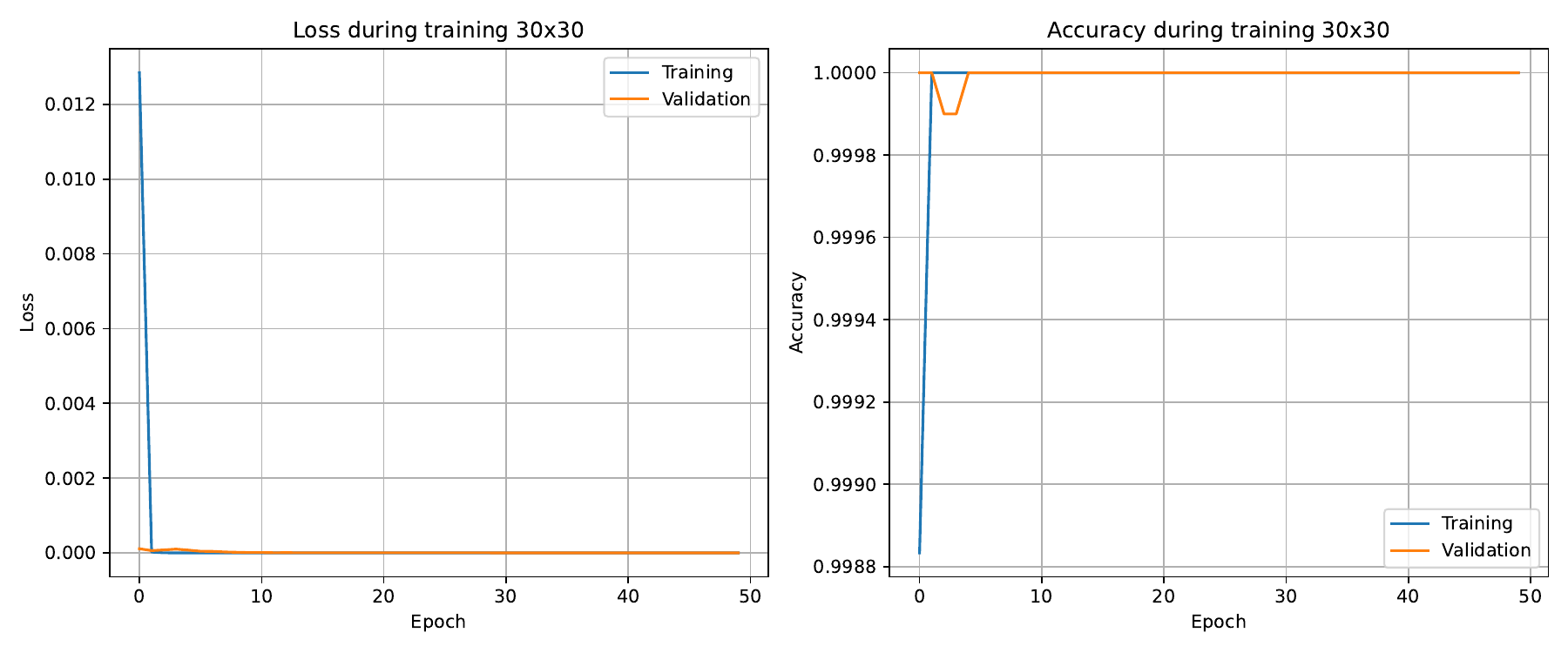}
    \caption{Accuracy and loss curves during training of the FFN for $n=30$. The curves show both training and validation performance.}
    \label{fig:training_plot_30x30_raw}
\end{figure}

Figure~\ref{fig:training_plot_30x30_raw} shows the accuracy and loss curves for $n = 30$, the other matrix sizes exhibited similar training behaviour. The figure illustrates that both the training and validation accuracy rapidly reach perfect classification, while the loss drops to near zero very early in the training process and remains stable for the remaining epochs. This behavior indicates that the neural network quickly identified the patterns in the data, achieving full separation between TP and non-TP matrices with minimal training time.

\subsection{Derived Features}
Instead of providing the raw entries of the matrix directly to the neural network, we considered a set of derived features that capture relevant structural and spectral properties of the matrix. The goal of this approach is to reduce the input dimensionality while preserving information that is strongly related to total positivity.

The features used as inputs included the determinant of the entire matrix, the Fiedler value, the trace of the matrix, all eigenvalues and the coefficients of the characteristic polynomial. Here the Fiedler value denotes the second smallest eigenvalue (the algebraic connectivity) of the graph Laplacian $L = D - W$ associated with the matrix, where $W=|A|$ is regarded as a weighted adjacency matrix and $D$ is the diagonal matrix of its row sums. It is therefore a spectral-graph connectivity descriptor of the matrix, distinct from the spectrum of $A$ itself. These features capture both the global and the spectral properties of the matrices, complementing the information contained in the raw entries. For each matrix size, the dataset was constructed by computing these features for all Totally Positive (TP) and Non-Totally Positive (Non-TP) examples. As in the previous subsection, TP matrices were generated through the product of bidiagonal pairs, ensuring diversity while preserving total positivity. The same feedforward neural network architecture and training configuration described in the previous subsection were used for this experiment.

\begin{figure}[tbp]
    \centering
    \includegraphics[width=1.0\textwidth]{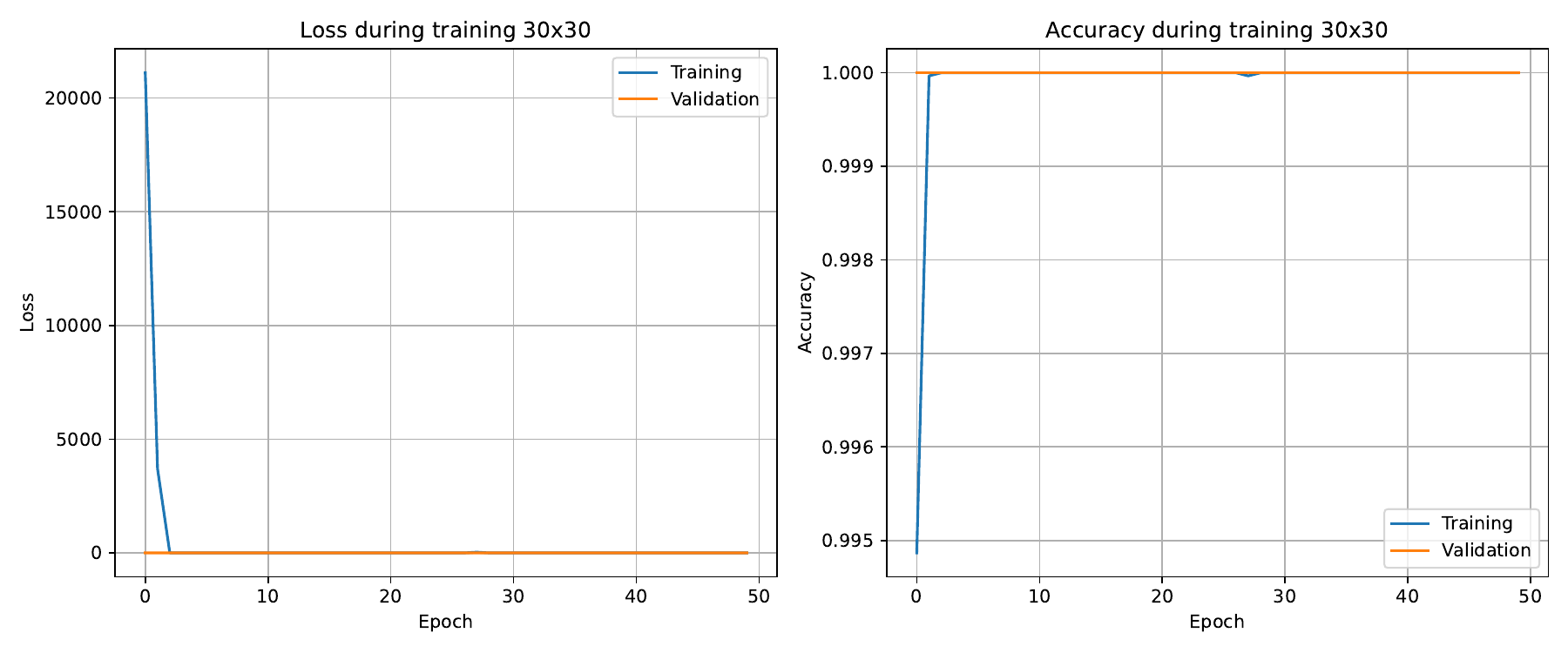}
    \caption{Training and validation accuracy and loss curves for the FFN trained on derived features for $n=30$.}
    \label{fig:acc_loss_derived}
\end{figure}

The training and validation accuracy and loss curves for $n=30$ are shown in Figure~\ref{fig:acc_loss_derived}, which is representative of all other matrix sizes considered. Similar to the networks trained on raw matrix entries, both the accuracy and loss curves reach high performance very quickly. Accuracy rapidly approaches its maximum value, while the loss drops sharply and stabilizes early in training, indicating that the networks efficiently captures the patterns present in the derived features with minimal training epochs.

After training the models on both raw matrix entries and derived features, the final test accuracies were identical across all considered matrix sizes ($5\times5$, $10\times10$ and $30\times30$), each achieving a perfect accuracy of 1.0. This result indicates that the derived features are equally effective at capturing the relevant structural information needed to identify total positivity, while offering a substantially more compact and interpretable representation compared to using all raw entries.

\section{Identification of the Most Relevant Coefficients}
\label{sec:feature_importance}
A central goal of Explainable Artificial Intelligence (XAI) is to uncover which input factors most influence the predictions of a model. In the context of our study, this corresponds to determining which mathematical features are most indicative of total positivity.
Feature importance analysis is therefore a core component of XAI, as it translates complex internal model representations into human-interpretable insights about the data. By identifying which derived features (such as the determinant, trace, or eigenvalues) contribute most to the model’s decision, it is possible to gain a deeper understanding of what properties distinguish totally positive matrices from others.

For that purpose, we applied two interpretability approaches: Integrated Gradients and SHAP. Each method provides a different perspective on the relationship between the features and the model’s predictions: Integrated Gradients attribute importance based on the gradients of the network, SHAP derives feature contributions using cooperative game theory.

For readers less familiar with these tools, we recall the underlying ideas, which are elementary. A \emph{feature attribution} assigns to each input feature a real number quantifying its influence on the model's output for a given prediction. Integrated Gradients~\cite{sundararajan2017} obtains this number by integrating the gradient of the model output along the straight segment joining a chosen baseline to the input (Eq.~\eqref{eq:ig} below), an exact path-integral identity that distributes the change in output among the input coordinates. SHAP values~\cite{lundberg2017}, by contrast, treat the features as players in a cooperative game and assign to each feature its Shapley value, that is, its average marginal contribution to the prediction taken over all orderings in which the features may be introduced. Both methods are used here only as exploratory devices to rank the spectral quantities; no conclusion of the paper depends on the internal details of either.

\subsection{Integrated Gradients}

Integrated Gradients (IG) is a gradient-based method for attributing feature importance in differentiable models such as neural networks~\cite{sundararajan2017}. For a given input \( x \) and a baseline \( x' \) (often chosen as a zero vector or an average input), IG computes the importance of each feature by integrating the gradients of the model’s output along the straight path between \( x' \) and \( x \):
\begin{equation}\label{eq:ig}
\text{IG}_i(x, x') = (x_i - x'_i) \int_0^1 \frac{\partial F(x' + \alpha (x - x'))}{\partial x_i} \, d\alpha
\end{equation}
where \( F(x) \) is the model’s prediction function.

This approach provides an attribution score for each input feature, reflecting how changes in that feature influence the output of the model. In this case, IG allows determination of how sensitive the classification of total positivity is to specific derived features such as the determinant, Fiedler value, eigenvalues or coefficients of the characteristic polynomial. By averaging attributions over the dataset, a global ranking of feature relevance is obtained, helping to identify which mathematical quantities are most critical for distinguishing TP matrices.

In this work, IG was applied to the matrix dimensions \(5,10\), and \(30\), using initially the same set of derived features that were used to train the classification models, and then a subset. A zero vector was chosen as the baseline.

\begin{figure}[tbp]
    \centering

    \begin{subfigure}{0.8\textwidth}
        \centering
        \IfFileExists{ig_5x5.pdf}
            {\includegraphics[width=\textwidth]{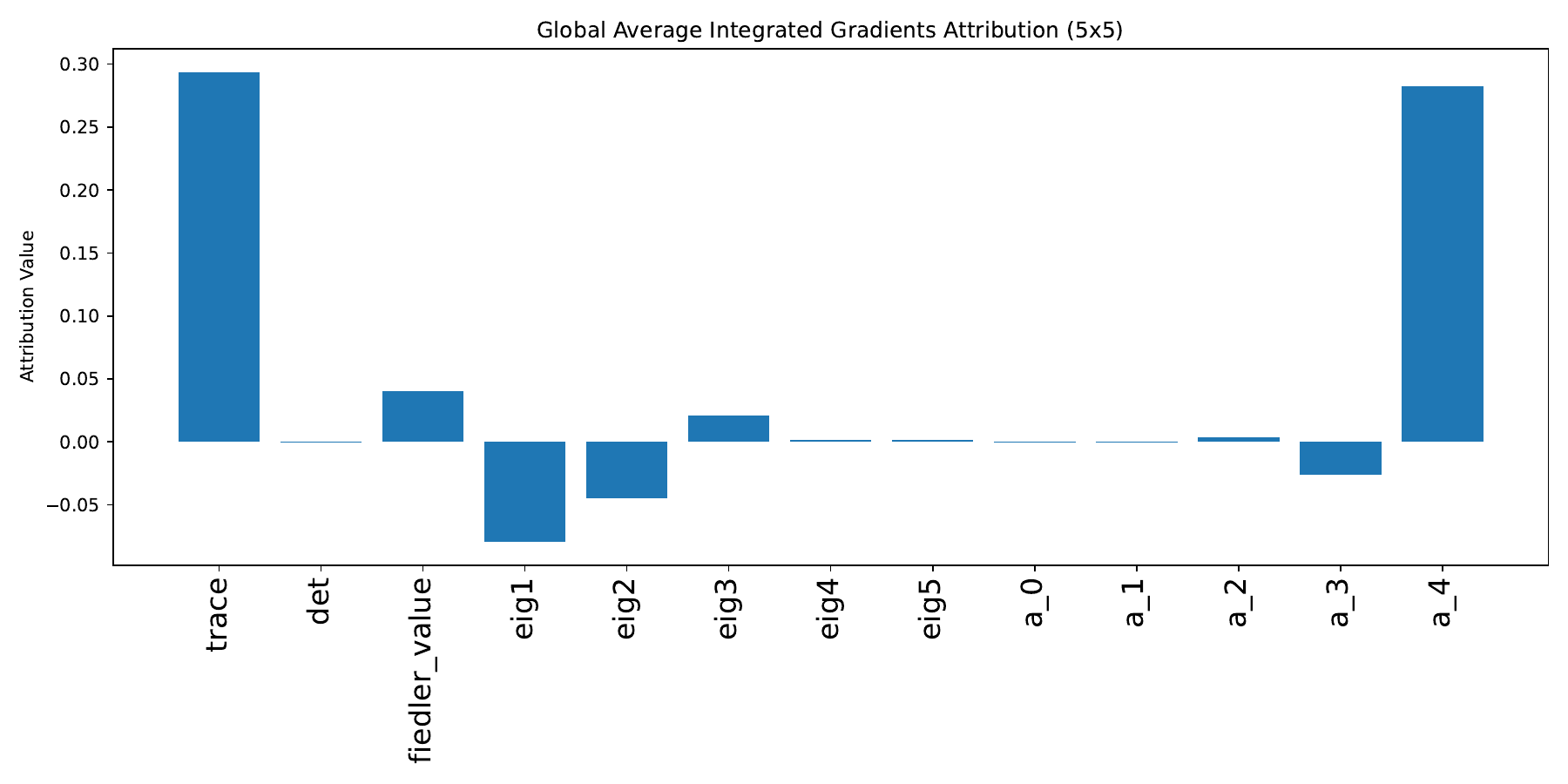}}
            {\fbox{\parbox[c][4cm][c]{\textwidth}{\centering Missing image}}}
        \caption{$n=5$.}
        \label{fig:ig_5x5}
    \end{subfigure}

    \begin{subfigure}{0.8\textwidth}
        \centering
        \IfFileExists{ig_10x10.pdf}
            {\includegraphics[width=\textwidth]{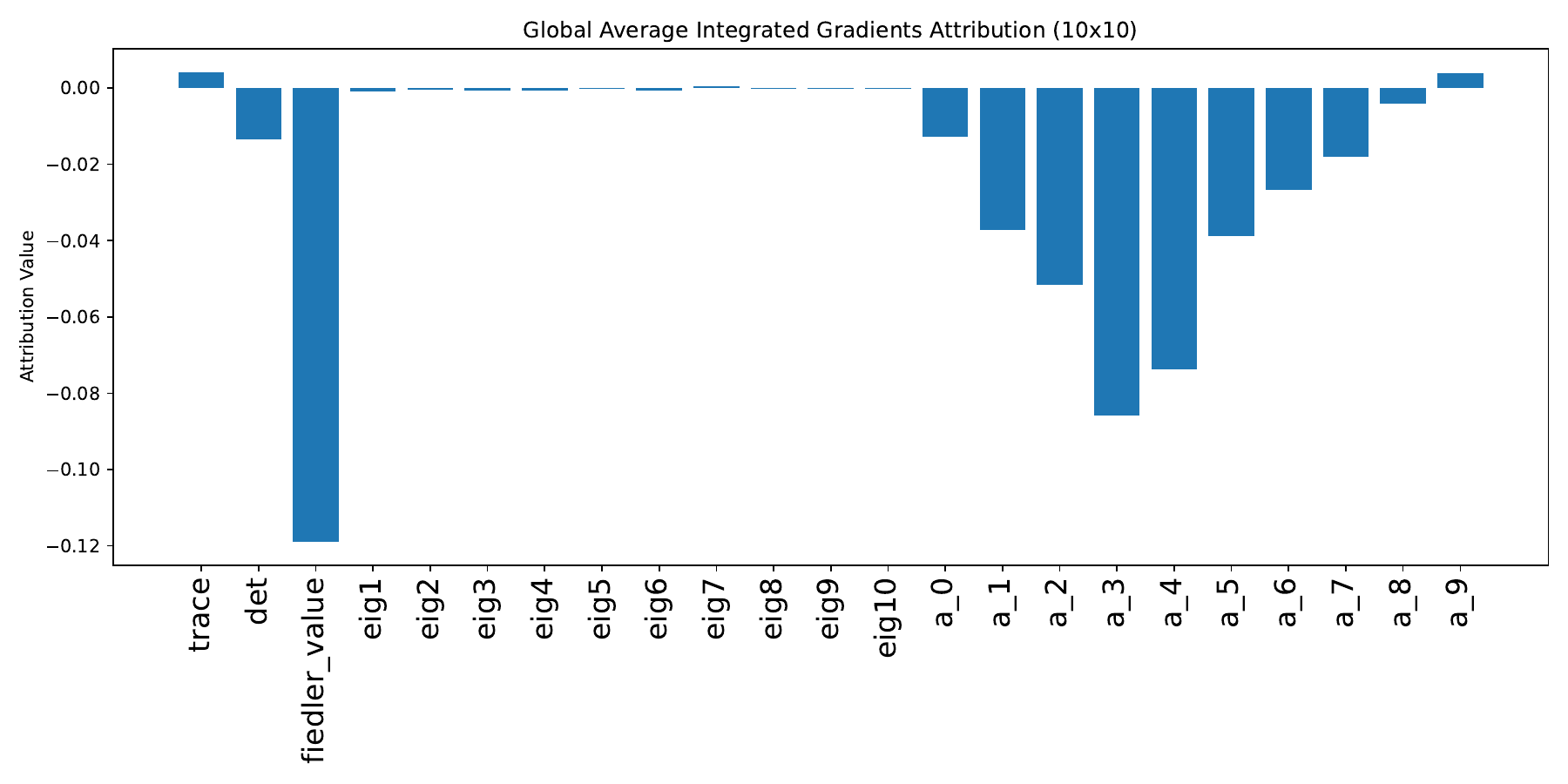}}
            {\fbox{\parbox[c][4cm][c]{\textwidth}{\centering Missing image}}}
        \caption{$n=10$.}
        \label{fig:ig_10x10}
    \end{subfigure}

    \begin{subfigure}{1.0\textwidth}
        \centering
        \IfFileExists{ig_30x30.pdf}
            {\includegraphics[width=\textwidth]{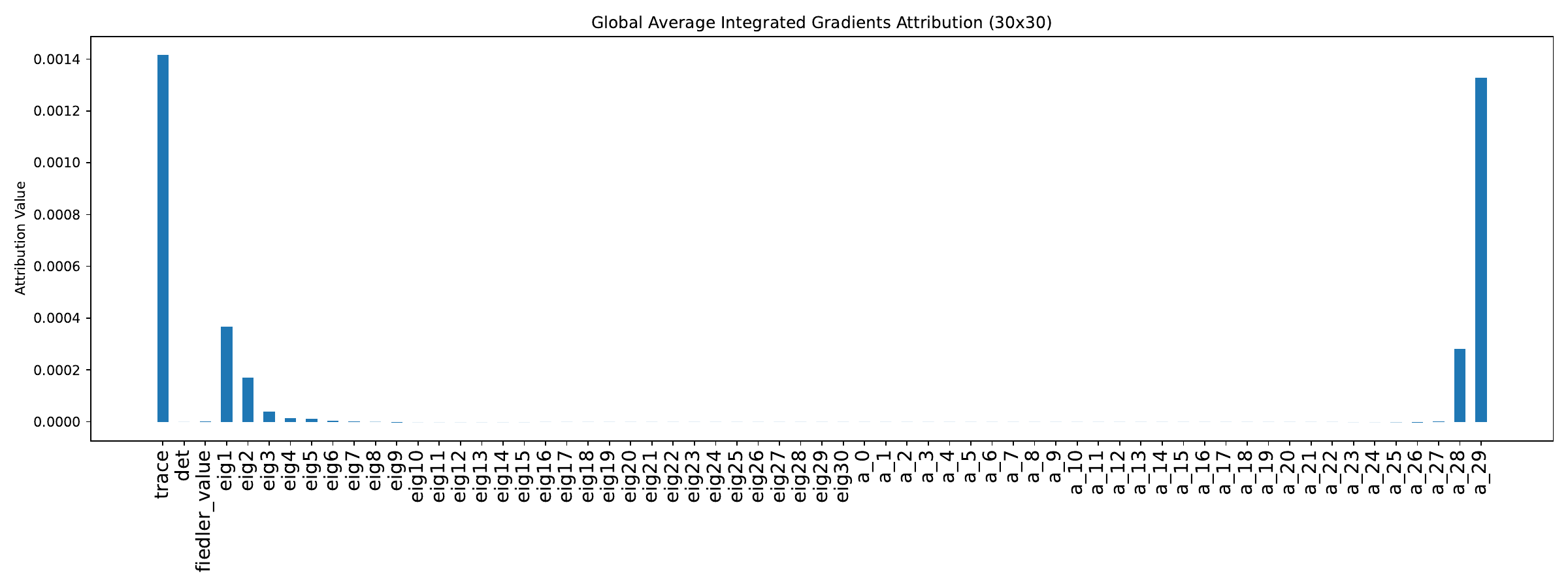}}
            {\fbox{\parbox[c][4cm][c]{\textwidth}{\centering Missing image}}}
        \caption{$n=30$.}
        \label{fig:ig_30x30}
    \end{subfigure}

    \caption{Integrated Gradients attributions for three matrix sizes:
    (a) $n=5$, (b) $n=10$, and (c) $n=30$.}
    \label{fig:ig_all}
\end{figure}

The IG attributions exhibit slight variations between runs due to stochastic effects in the model and sampling. However, as shown in Figure~\ref{fig:ig_all}, a consistent pattern emerges in all cases: the coefficients of the characteristic polynomial consistently show high importance, particularly those of higher orders. For the $30 \times 30$ case, the trace of the matrix achieved a very high score, which is reasonable since  the trace is the negative of the highest-order coefficient of the characteristic polynomial ($a_{n-1} = -tr(A)$). This indicates that these coefficients capture key structural information relevant to distinguishing totally positive matrices from non–totally positive ones.

The Fiedler value also shows high importance across runs, suggesting that it captures meaningful connectivity information. We emphasize that this is not a distinguished eigenvalue of $A$: since a TP matrix already has simple positive eigenvalues, no individual eigenvalue of $A$ is structurally singled out, which is precisely why we included a descriptor drawn from a different object, the graph Laplacian. The Fiedler value was thus one of several derived features considered in this exploratory analysis; nevertheless, it will not be considered further in this study, as it is a single scalar feature and does not offer sufficient dimensionality to capture the individual contributions of multiple structural components of the matrix.

To further investigate this pattern, IG was then applied using only the characteristic polynomial coefficients as input features, for $n=5,10,30$. This analysis isolates the contribution of these coefficients without the influence of other derived features.

\begin{figure}[tbp]
    \centering

    \begin{subfigure}{0.8\textwidth}
        \centering
        \IfFileExists{ig_5x5_charpoly.pdf}
            {\includegraphics[width=\textwidth]{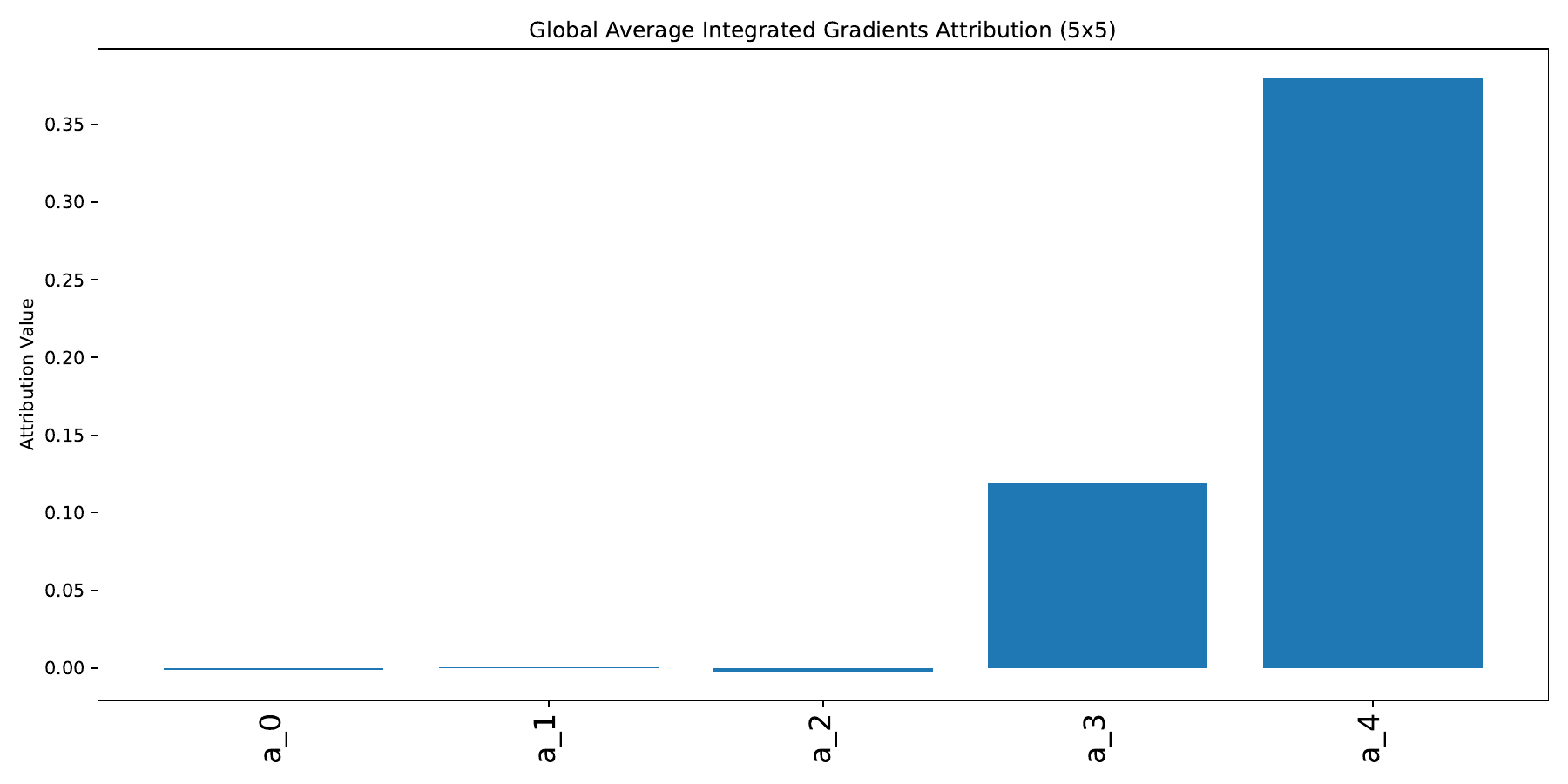}}
            {\fbox{\parbox[c][4cm][c]{\textwidth}{\centering Missing image}}}
        \caption{$n = 5$}
        \label{fig:ig_charpolycoeffs_5x5}
    \end{subfigure}

    \begin{subfigure}{0.8\textwidth}
        \centering
        \IfFileExists{ig_10x10_charpoly.pdf}
            {\includegraphics[width=\textwidth]{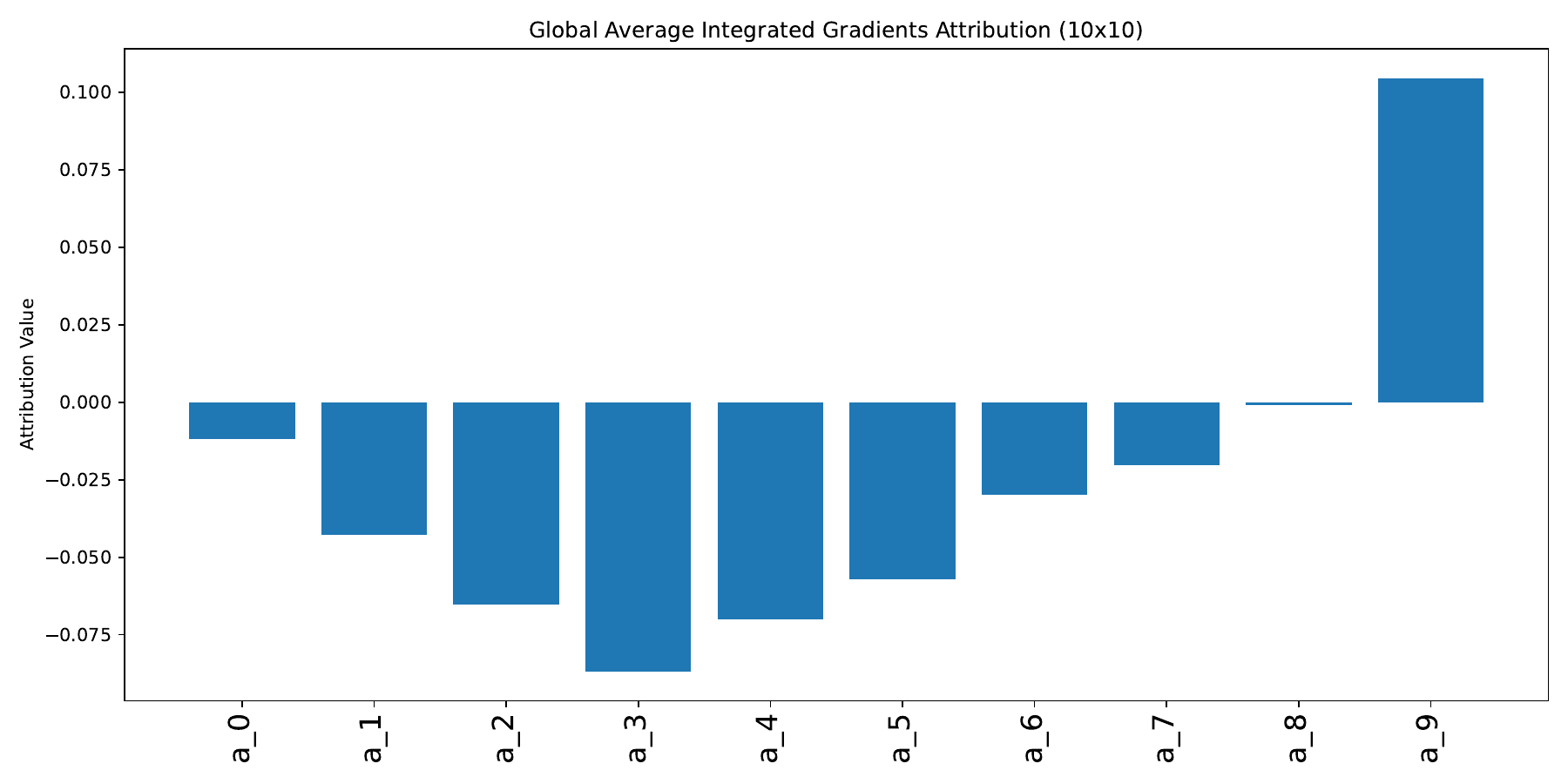}}
            {\fbox{\parbox[c][4cm][c]{\textwidth}{\centering Missing image}}}
        \caption{$n = 10$}
        \label{fig:ig_charpolycoeff_10x10}
    \end{subfigure}

    \begin{subfigure}{0.8\textwidth}
        \centering
        \IfFileExists{ig_30x30_charpoly.pdf}
            {\includegraphics[width=\textwidth]{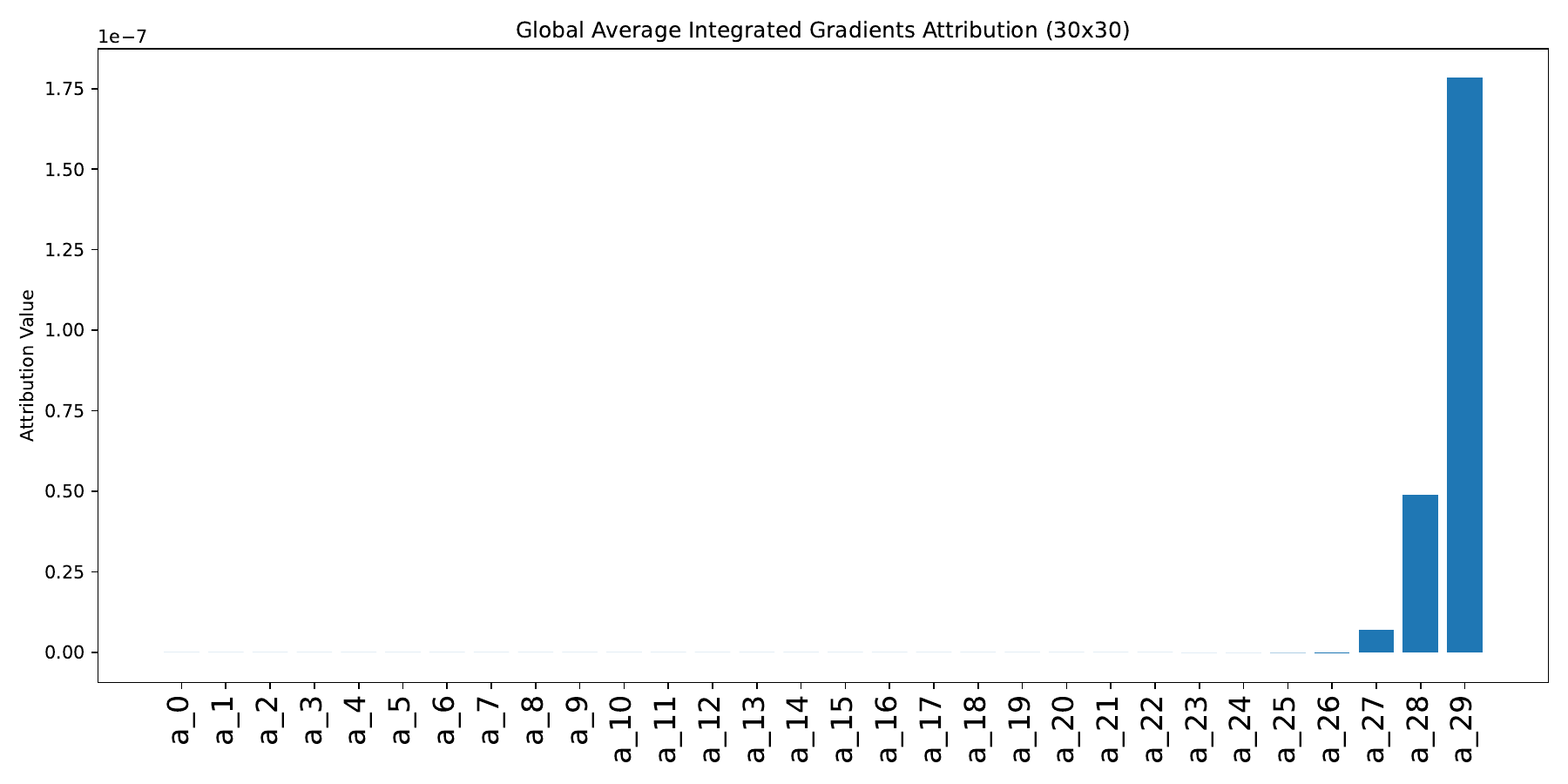}}
            {\fbox{\parbox[c][4cm][c]{\textwidth}{\centering Missing image}}}
        \caption{$n = 30$}
        \label{fig:ig_charpolycoeff_30x30}
    \end{subfigure}

    \caption{Integrated Gradients attributions using only characteristic polynomial coefficients for (a) $n=5$, (b) $n=10$, and (c) $n=30$.}
    \label{fig:ig_charpoly_all}
\end{figure}

Apart from the \(10\times10\) case, which shows a more balanced distribution of importance among the coefficients (see Figure~\ref{fig:ig_charpolycoeff_10x10}), the other cases clearly display a tendency for the higher-order coefficients to be more important (Figures~\ref{fig:ig_charpolycoeffs_5x5} and \ref{fig:ig_charpolycoeff_30x30}). We interpret the $10\times10$ behaviour as a transitional, finite-size effect: $n=5$ is a small case with few coefficients, $n=30$ is large enough for the higher-order coefficients to dominate clearly, and $n=10$ lies in an intermediate regime in which this ordering has not yet emerged. This reading is also consistent with the run-to-run variability of the attributions noted above. In any case it does not affect our conclusions, since the model restricted to the three highest-order coefficients classifies the $10\times10$ matrices with only two errors in $10^4$ test examples (Section~\ref{sec:restricted_coeffs}). The balance of the attributions concerns only which coefficients a model trained on all of them relies on; whether the three highest-order coefficients suffice on their own is a separate question, settled directly by that classifier. Overall, this pattern reinforces the idea that the higher-order coefficients of the characteristic polynomial capture key structural information relevant to distinguishing totally positive matrices from non-totally positive ones, while the lower-order coefficients play a comparatively smaller role, especially for larger matrices such as the \(30\times30\) case.

Finally, IG was computed again using only the three highest–order characteristic polynomial coefficients for each matrix size. This focused analysis aims to determine whether the most dominant coefficients alone can explain the classification behavior observed previously.

\begin{figure}[tbp]
    \centering

    \begin{subfigure}{0.8\textwidth}
        \centering
        \IfFileExists{ig_charpoly_top3_5x5.pdf}
            {\includegraphics[width=\textwidth]{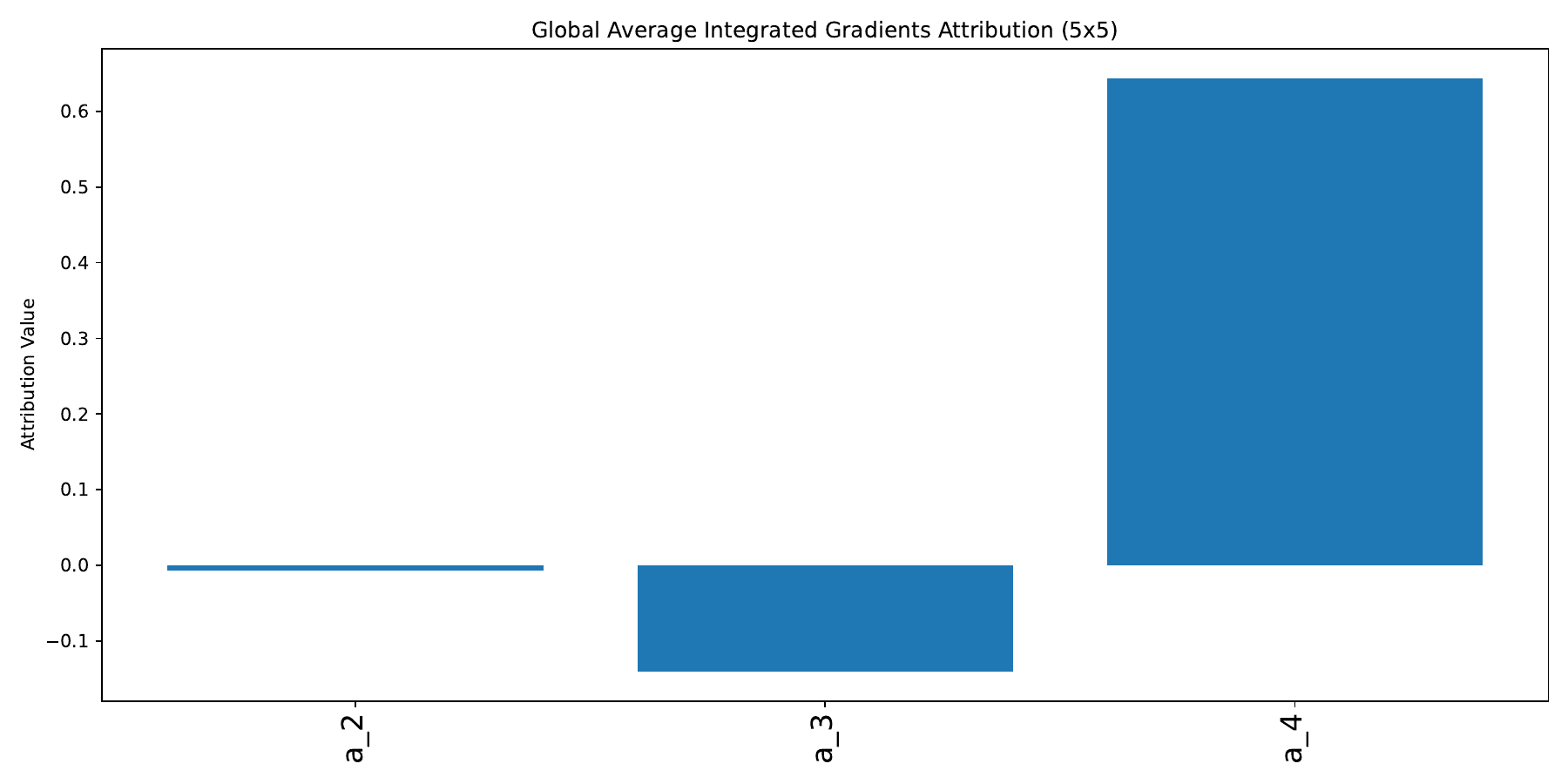}}
            {\fbox{\parbox[c][4cm][c]{\textwidth}{\centering Missing image}}}
        \caption{$n = 5$}
        \label{fig:ig_charpoly_top3_5x5}
    \end{subfigure}

    \begin{subfigure}{0.8\textwidth}
        \centering
        \IfFileExists{ig_charpoly_top3_10x10.pdf}
            {\includegraphics[width=\textwidth]{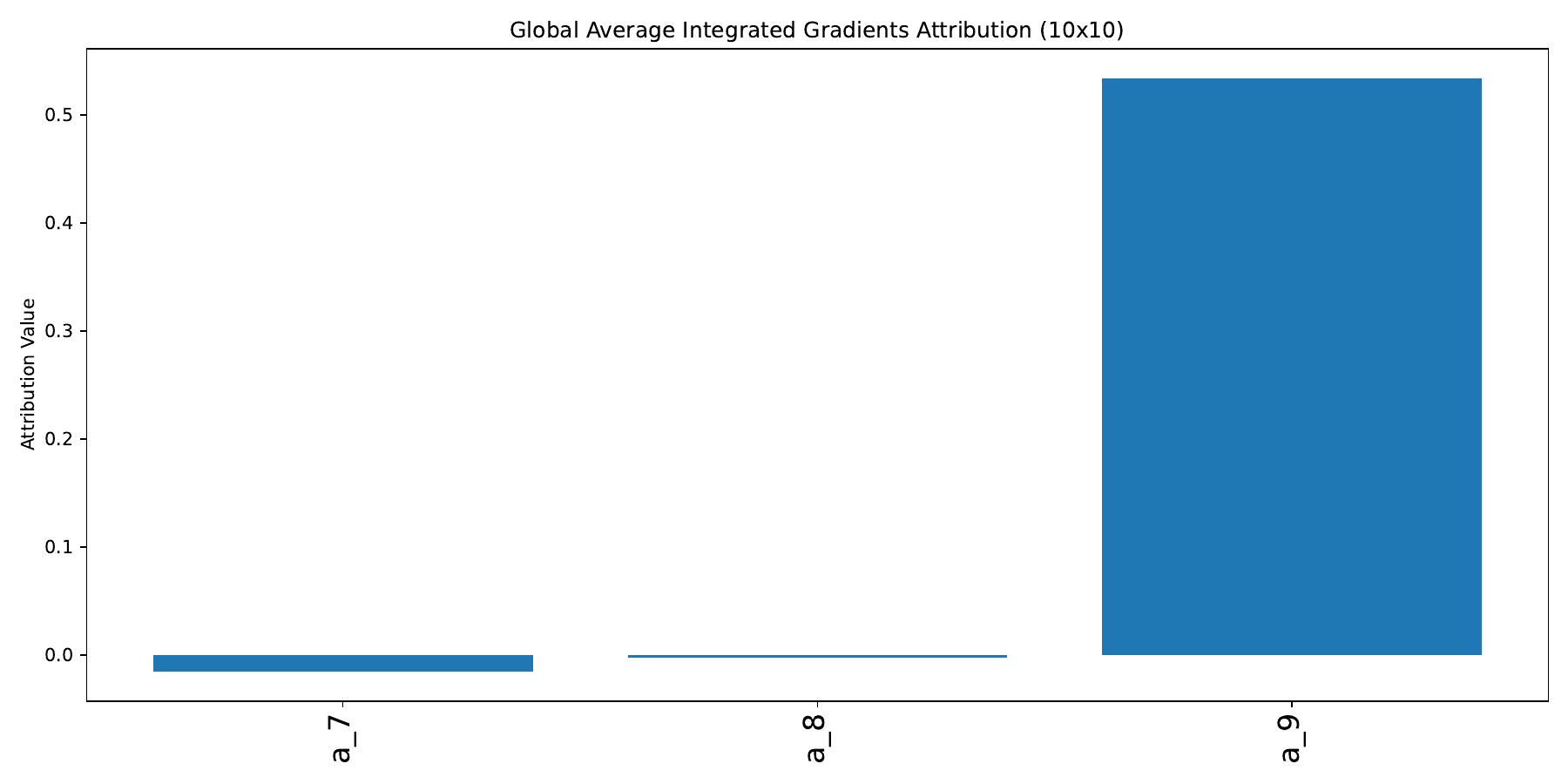}}
            {\fbox{\parbox[c][4cm][c]{\textwidth}{\centering Missing image}}}
        \caption{$n = 10$}
        \label{fig:ig_charpoly_top3_10x10}
    \end{subfigure}

    \begin{subfigure}{0.8\textwidth}
        \centering
        \IfFileExists{ig_charpoly_top3_30x30.pdf}
            {\includegraphics[width=\textwidth]{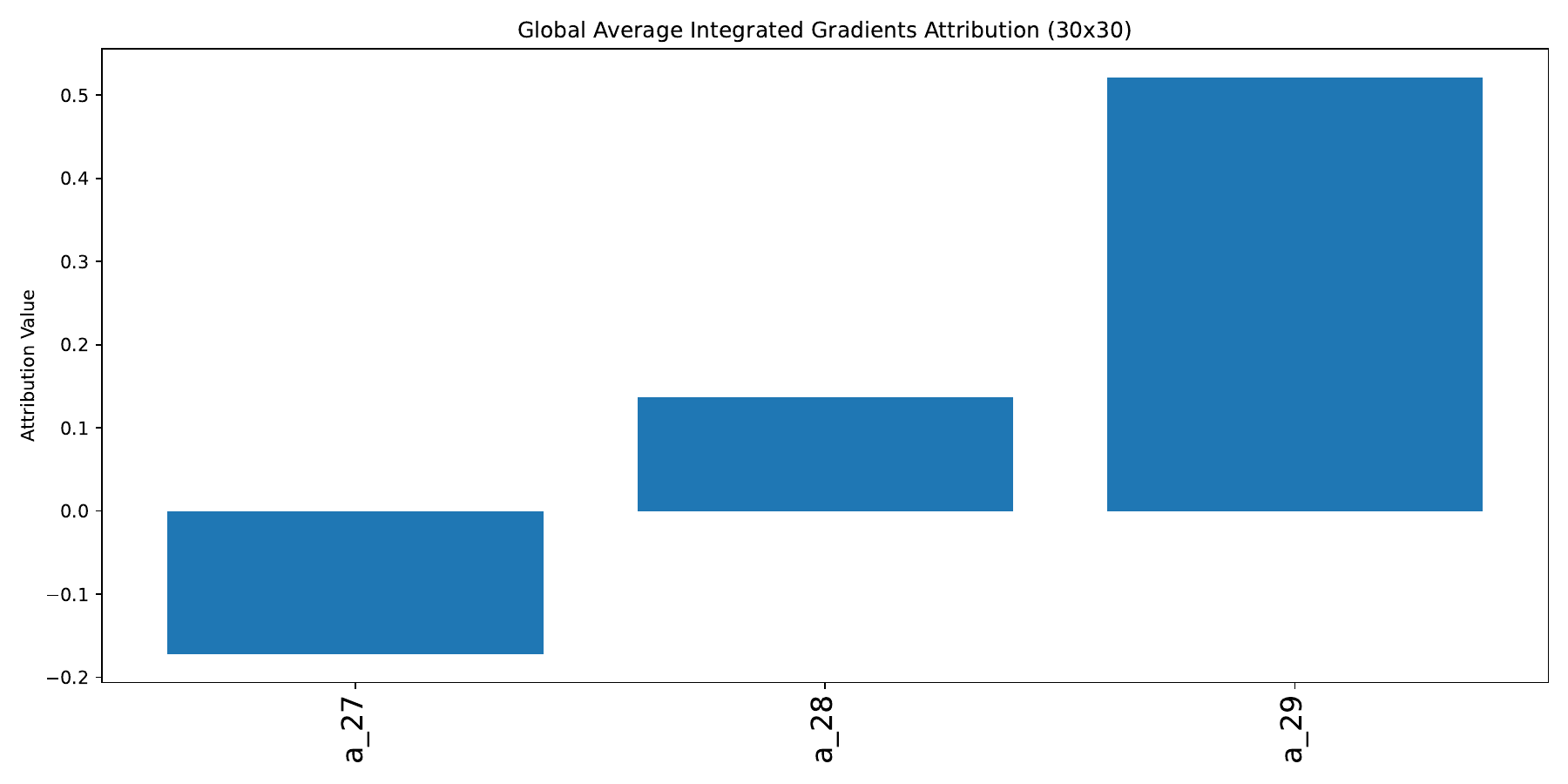}}
            {\fbox{\parbox[c][4cm][c]{\textwidth}{\centering Missing image}}}
        \caption{$n = 30$}
        \label{fig:ig_charpoly_top3_30x30}
    \end{subfigure}

    \caption{Integrated Gradients attributions using the three highest–order characteristic polynomial coefficients for (a) $n = 5$, (b) $n = 10$, and (c) $n = 30$.}
    \label{fig:ig_charpoly_top3_all}
\end{figure}

The results of this focused analysis, shown in Figure~\ref{fig:ig_charpoly_top3_all}, confirm the dominant role of the highest-order characteristic polynomial coefficient in the classification task. For all matrix sizes, the highest-order coefficient consistently exhibits the largest IG attribution, while the contributions of the next two lower-order coefficients are smaller.

Across all experiments, a clear trend is observed: although the exact attribution values fluctuate slightly between runs, the higher–degree characteristic polynomial coefficients consistently dominate the importance ranking. This confirms their key role in capturing the structural properties that determine total positivity, even when considered in isolation from other derived features.

\subsection{SHAP Values}

SHAP is a method for explaining model predictions based on concepts from cooperative game theory~\cite{lundberg2017}. It assigns each feature a value that represents its contribution to the model’s output, by comparing the model’s predictions with and without that feature. In essence, SHAP values quantify how much each feature helps (or hinders) the prediction when considered in combination with others.

While the exact computation of SHAP values is computationally intensive, efficient approximation methods allow their use even in complex models such as neural networks. In this work, SHAP was applied to the \(5\times5\), \(10\times10\), and \(30\times30\) matrix cases using the same derived features as in the previous analyses.

\begin{figure}[tbp]
    \centering

    \begin{subfigure}{0.45\textwidth}
        \centering
        \IfFileExists{shap_5x5.pdf}
            {\includegraphics[width=\textwidth]{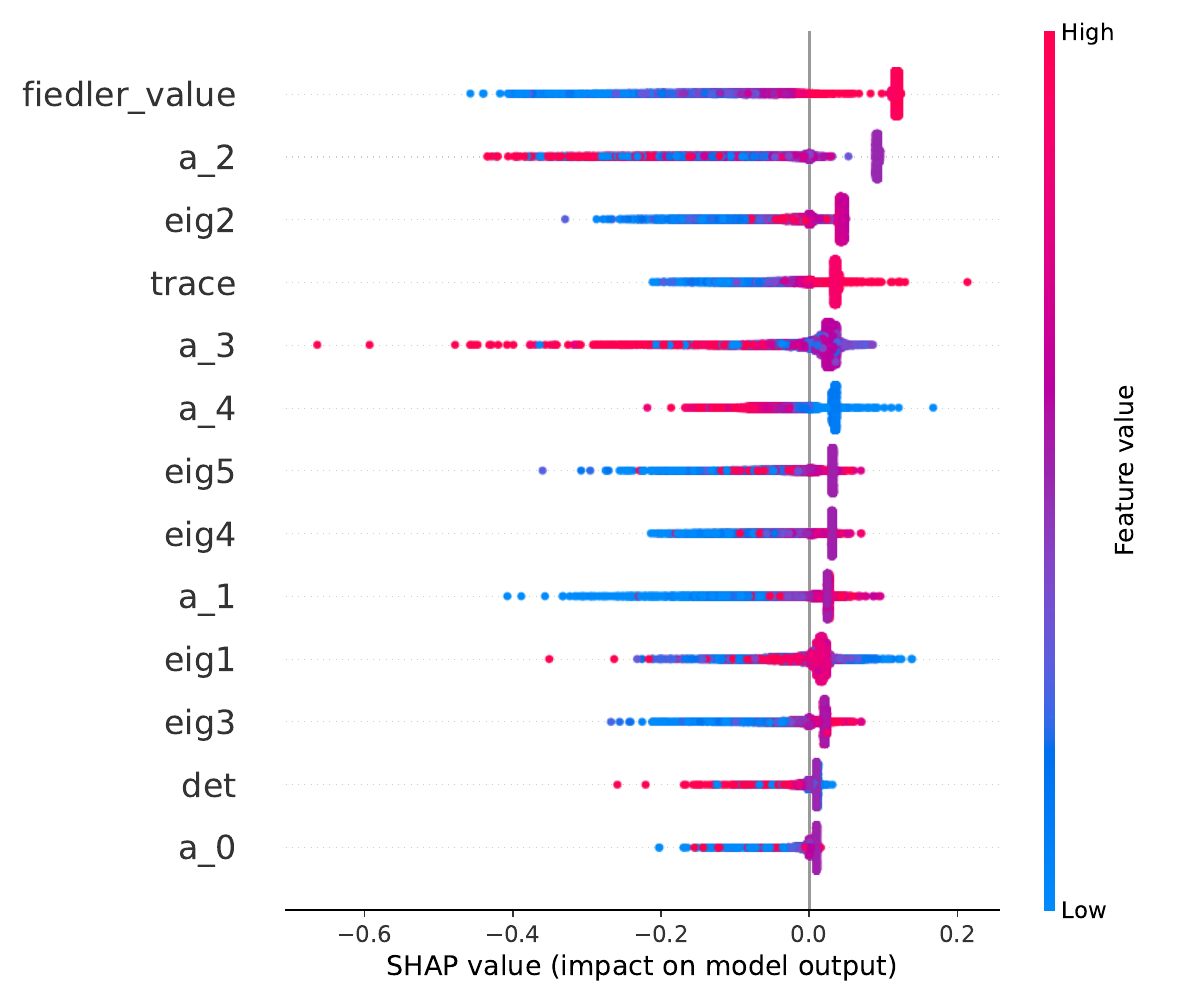}}
            {\fbox{\parbox[c][4cm][c]{\textwidth}{\centering Placeholder for missing image}}}
        \caption{5×5 case}
        \label{fig:shap_5x5}
    \end{subfigure}
    \hfill
    \begin{subfigure}{0.45\textwidth}
        \centering
        \IfFileExists{shap_10x10.pdf}
            {\includegraphics[width=\textwidth]{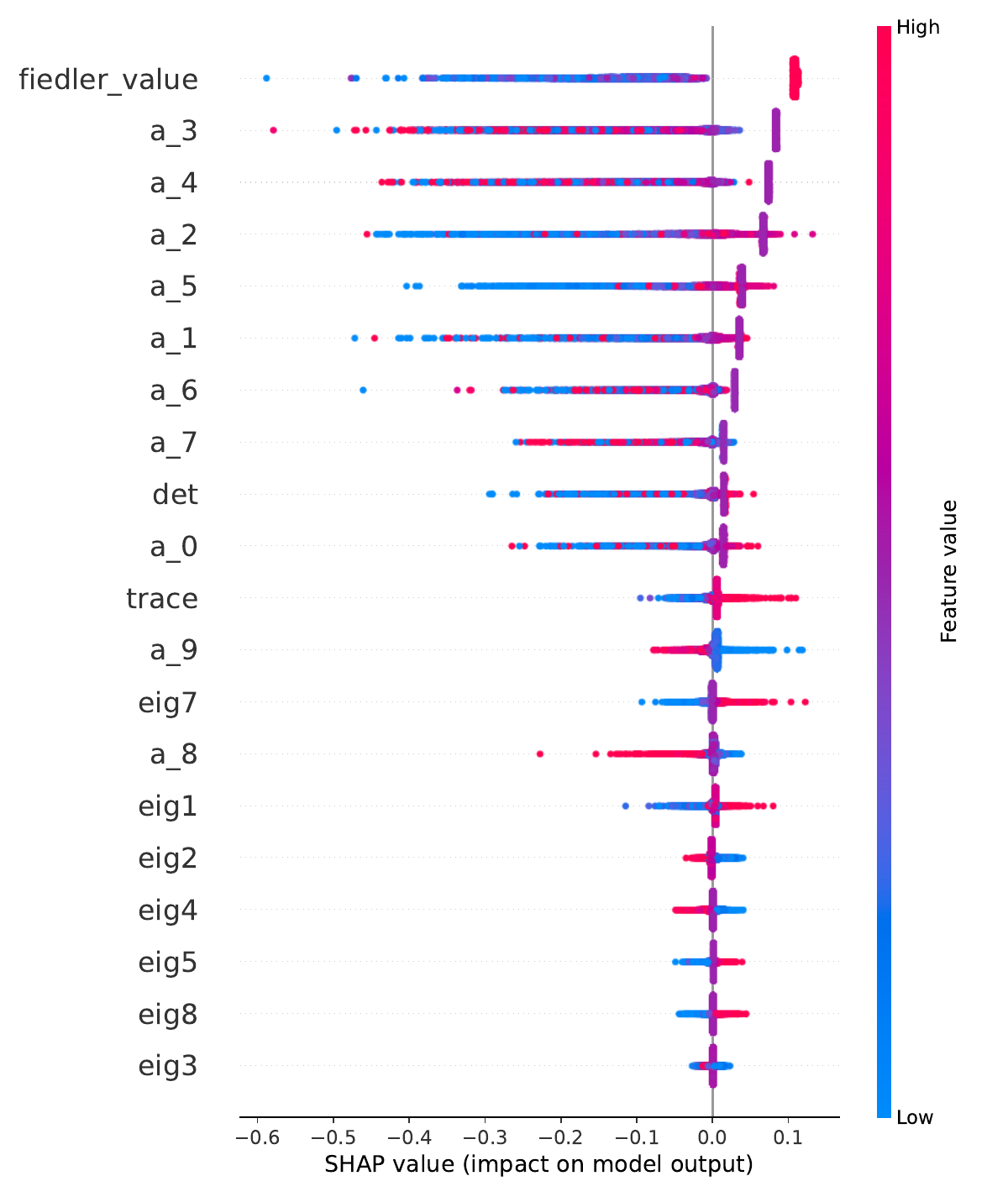}}
            {\fbox{\parbox[c][4cm][c]{\textwidth}{\centering Placeholder for missing image}}}
        \caption{10×10 case}
        \label{fig:shap_10x10}
    \end{subfigure}
    \hfill
    \begin{subfigure}{0.45\textwidth}
        \centering
        \IfFileExists{shap_30x30.pdf}
            {\includegraphics[width=\textwidth]{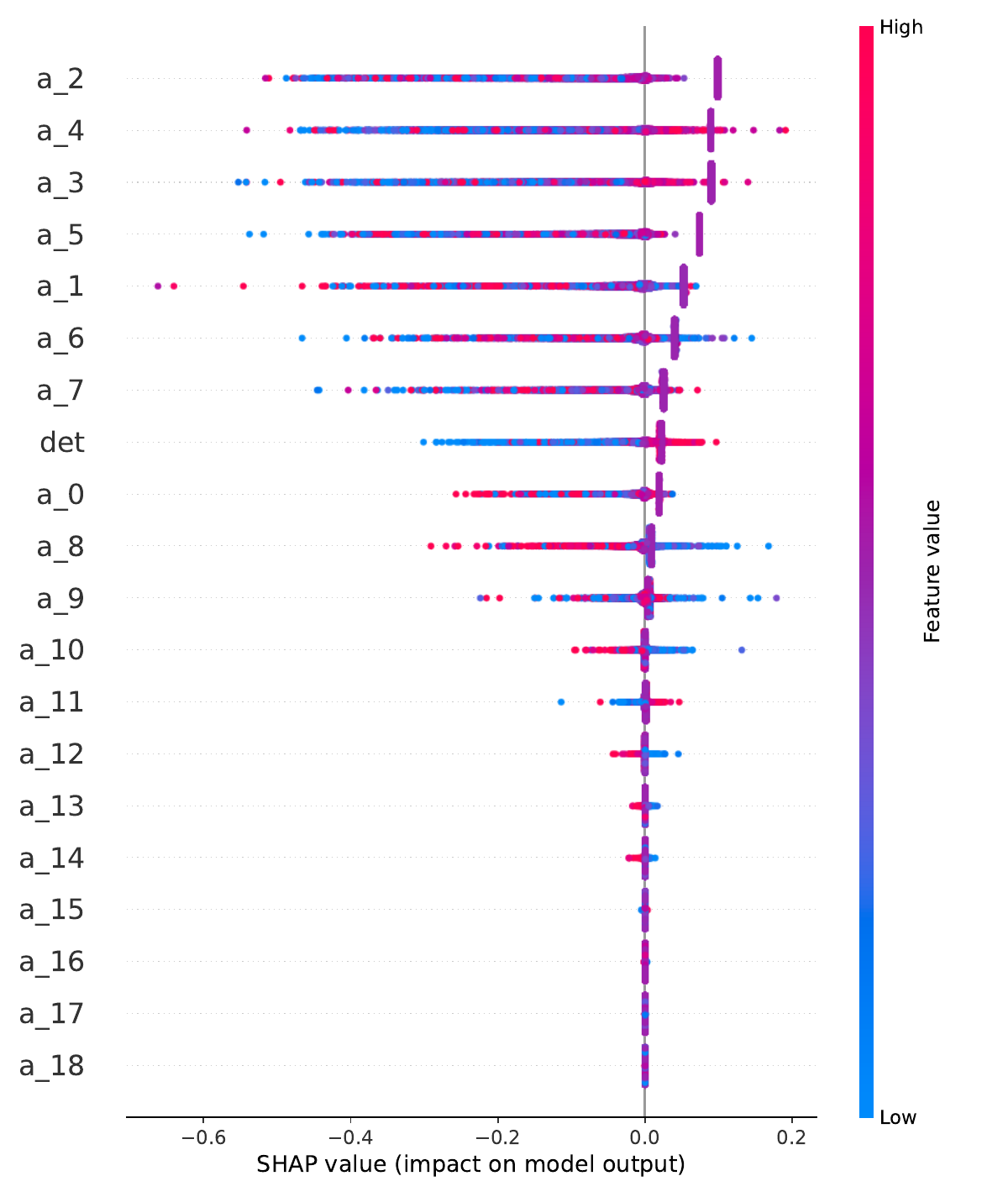}}
            {\fbox{\parbox[c][4cm][c]{\textwidth}{\centering Placeholder for missing image}}}
        \caption{30×30 case}
        \label{fig:shap_30x30}
    \end{subfigure}

    \caption{SHAP feature importance across matrix sizes.
    (a) 5×5, (b) 10×10, and (c) 30×30.}
    \label{fig:shap_all}
\end{figure}

Figure~\ref{fig:shap_all} presents the SHAP feature importance for the derived features across the three matrix sizes considered. For the $5\times5$ case (Figure~\ref{fig:shap_5x5}), the attributions are relatively balanced between features, with the Fiedler value exhibiting the highest importance. A clear separation is observed between high (red) and low (blue) feature values: higher feature values (red) contribute positively to classification as TP (class 1), whereas lower values (blue) contribute negatively (class 0).

In the $10\times10$ case (Figure~\ref{fig:shap_10x10}), the pattern is similar, although the characteristic polynomial coefficients begin to emerge as more influential than most features, with the exception of the Fiedler value, which remains highly ranked. This indicates that as the matrix size increases, structural information captured by the coefficients of the characteristic polynomial becomes increasingly relevant

For the $30\times30$ case (Figure~\ref{fig:shap_30x30}), the Fiedler value is no longer prominent, consistent with the observations from the IG analysis. The coefficients of the characteristic polynomial dominate the SHAP ranking, suggesting that for larger matrices, these coefficients encode the majority of the information necessary for identifying total positivity. Overall, the SHAP analysis reinforces the importance of the characteristic polynomial as a compact and informative representation of the matrix.

\begin{figure}[tbp]
    \centering

    \begin{subfigure}{0.5\textwidth}
        \centering
        \IfFileExists{shap_5x5_coeffs.pdf}
            {\includegraphics[width=\textwidth]{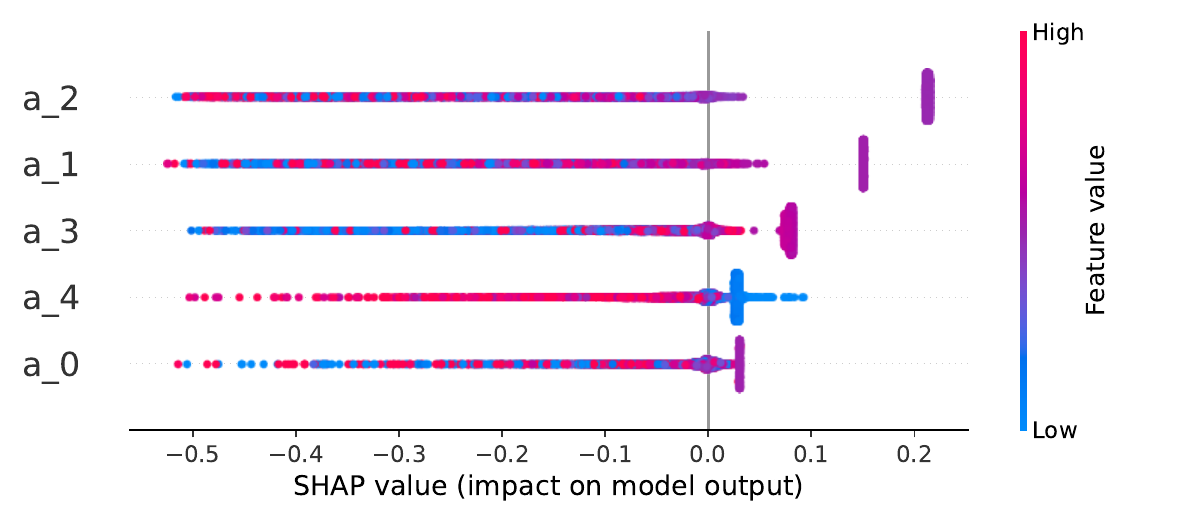}}
            {\fbox{\parbox[c][4cm][c]{\textwidth}{\centering Placeholder for missing image}}}
        \caption{5×5 case}
        \label{fig:shap_charpoly_5x5}
    \end{subfigure}

    \begin{subfigure}{0.45\textwidth}
        \centering
        \IfFileExists{shap_10x10_coeffs.pdf}
            {\includegraphics[width=\textwidth]{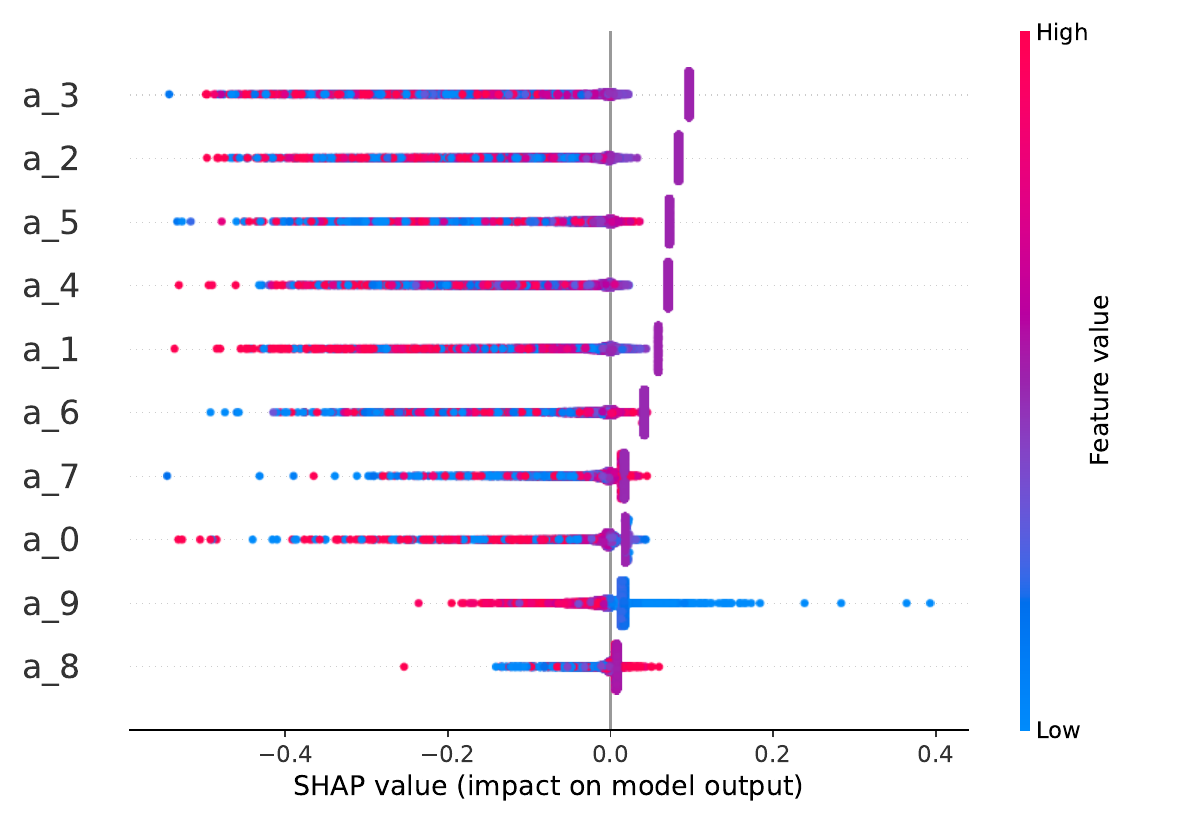}}
            {\fbox{\parbox[c][4cm][c]{\textwidth}{\centering Placeholder for missing image}}}
        \caption{10×10 case}
        \label{fig:shap_charpoly_10x10}
    \end{subfigure}
    \begin{subfigure}{0.45\textwidth}
        \centering
        \IfFileExists{shap_30x30_coeffs.pdf}
            {\includegraphics[width=\textwidth]{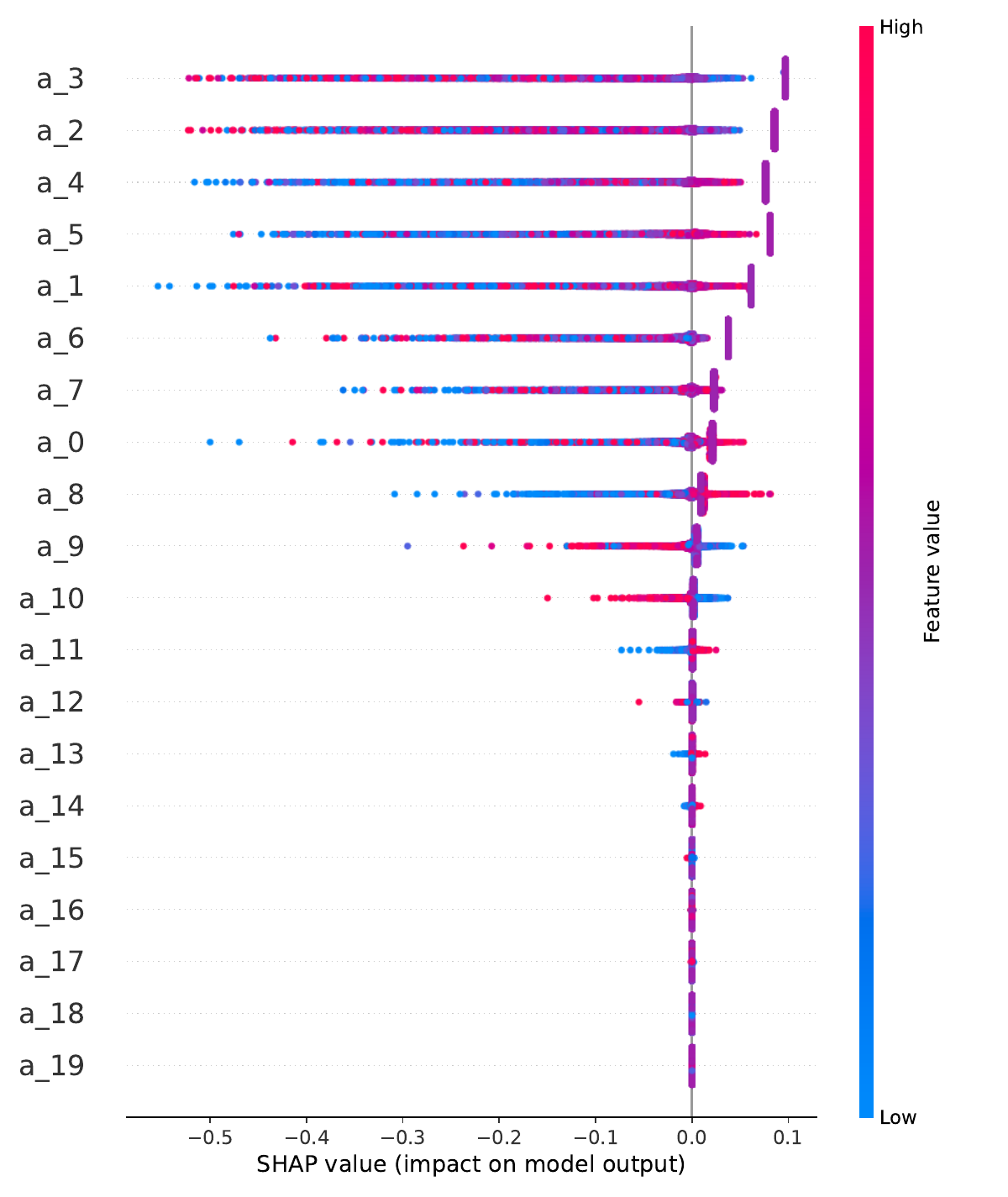}}
            {\fbox{\parbox[c][4cm][c]{\textwidth}{\centering Placeholder for missing image}}}
        \caption{30×30 case}
        \label{fig:shap_charpoly_30x30}
    \end{subfigure}

    \caption{SHAP values using only characteristic polynomial coefficients across matrix sizes:
    (a) 5×5, (b) 10×10, and (c) 30×30.}
    \label{fig:shap_charpoly_all}
\end{figure}

Figure~\ref{fig:shap_charpoly_all} shows the SHAP values when only the characteristic polynomial coefficients are used as input features, for the \(5\times5\), \(10\times10\) and \(30\times30\) matrices. Contrary to the IG analysis, which suggested that higher-order coefficients consistently have greater importance, the SHAP results indicate a more balanced contribution across all coefficients. No clear pattern emerges where the higher-order coefficients dominate, instead the importance is distributed relatively evenly among the coefficients for all matrix sizes.

This finding suggests that, while the characteristic polynomial coefficients collectively encode the information necessary for classification, SHAP attributes importance more uniformly across them, highlighting that the model may rely on multiple coefficients simultaneously rather than emphasizing higher-order terms. The results reinforce the complementarity of different interpretability methods, showing that IG and SHAP can provide distinct perspectives on the relevance of features.

\section{Restriction to the Three Highest-Order Characteristic Coefficients}
\label{sec:restricted_coeffs}

IG strongly prioritizes the highest-order coefficients; SHAP confirms that the characteristic polynomial coefficients (as a group) dominate. Motivated by IG and the excellent empirical performance, we restrict to the three highest-order coefficients and
new Feed forward Neural Networks (FFNs) were trained using only these coefficients as input features.

The goal was to evaluate how much of the predictive capacity of the full model could be retained when restricted to the most informative subset of features.

The network architecture and training parameters were kept identical to those used in the previous FFN experiments. This includes the same number of hidden layers, neurons per layer, activation functions, optimization algorithm and learning rate schedule. The only modification was the dimensionality of the input layer, reduced to three neurons corresponding to the selected coefficients.

Despite the drastic reduction in input dimensionality, the model still discriminated between the two classes almost perfectly. On a test set of $10^4$ matrices, using only the three coefficients $(a_{n-1},a_{n-2},a_{n-3})$, it attained an accuracy of $0.9998$ for $5\times5$ ($2$ misclassifications), $0.9998$ for $10\times10$ ($2$ misclassifications) and $0.9999$ for $30\times30$ ($1$ misclassification).
This demonstrates that a significant portion of the relevant structure is encoded in the higher-order coefficients of the characteristic polynomial, which are directly related to global invariants such as the trace and determinant.

The confusion matrices below summarize the results. For the binary TP/non-TP problem, the confusion matrix is the $2\times2$ array whose rows index the true class and whose columns index the predicted class, both ordered as $(\text{non-TP},\,\text{TP})$. Its top-left and bottom-right entries are therefore the numbers of correctly classified non-TP and TP matrices, while its top-right and bottom-left entries are the two types of error (non-TP predicted as TP, and TP predicted as non-TP). Notably, the overall classification accuracy remained comparable to that obtained with the full set of derived features, indicating that these three coefficients capture most of the relevant information for the TP discrimination problem.

\[
\text{Confusion Matrix } (5\times5):
\quad
\begin{bmatrix}
5034 & 1 \\
1    & 4964
\end{bmatrix}.
\]

\vspace{0.3cm}

\[
\text{Confusion Matrix } (10\times10):
\quad
\begin{bmatrix}
5059 & 2 \\
0    & 4939
\end{bmatrix}.
\]

\vspace{0.3cm}

\[
\text{Confusion Matrix } (30\times30):
\quad
\begin{bmatrix}
5007 & 1 \\
0    & 4992
\end{bmatrix}.
\]

\section{Linear and Nonlinear Separation in Coefficient Space}
\label{sec:svm}
To further investigate the structure of the feature space defined by the characteristic polynomial coefficients, Support Vector Machines (SVMs) were trained using only the three highest-order coefficients as input features. Two types of kernels were considered: a linear kernel and a radial basis function (RBF) kernel.

\subsection{Linear Kernel}

The linear SVM provides a direct geometric interpretation by attempting to construct a single hyperplane separating the TP and non-TP samples.
The resulting decision boundary can be expressed as:
\[
w_1 x_1 + w_2 x_2 + w_3 x_3 + b = 0,
\]
where \(w_1, w_2, w_3\) are the learned weights and \(b\) is the bias term. The equations obtained for the linear SVMs trained on matrices of different sizes are:
\[
\begin{aligned}
 n=5: \quad & -0.26 a_2 + 0.22 a_3 - 1.78 a_4 - 1.42 = 0 \\
n=10: \quad & -0.09 a_7 + 0.21 a_8 - 0.99 a_9 - 0.55 = 0 \\
n=30: \quad & -0.02 a_{27} + 0.09 a_{28} - 0.52 a_{29} - 0.01 = 0
\end{aligned}
\]

\begin{figure}[tbp]
    \centering  \IfFileExists{svm_hyperplanes_combined.pdf}
    {\includegraphics[width=0.8\textwidth]{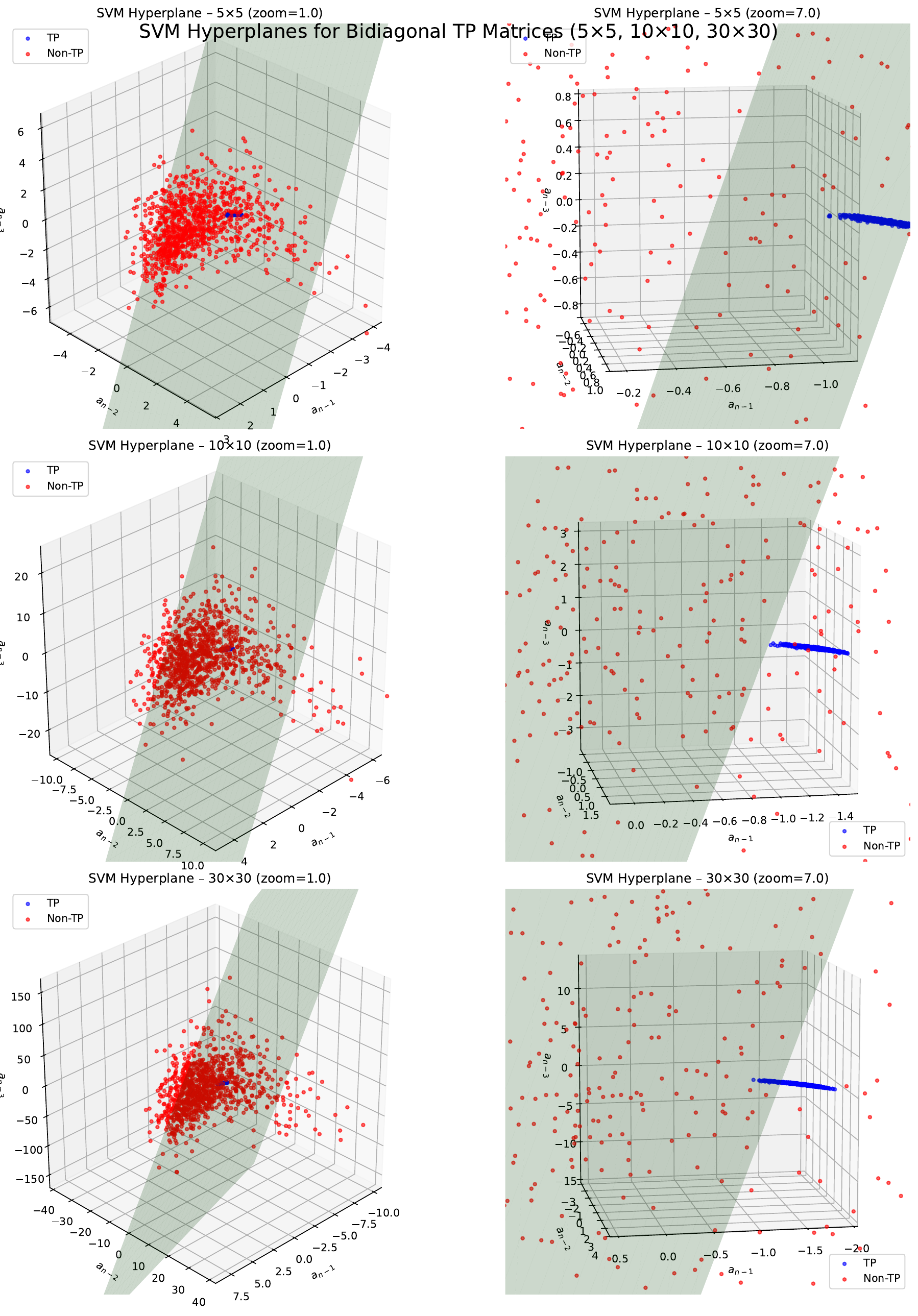}}
    {\fbox{\parbox[c][5cm][c]{0.8\textwidth}{\centering Placeholder for missing image}}}
    \caption{Separating Hyperplane in the Feature Space for SVM Classification of 5x5, 10x10 and 30x30 Totally Positive (TP) and Non-Totally Positive (Non-TP) Matrices}
    \label{fig:svm_hyperplanes_combined}
\end{figure}

Figure~\ref{fig:svm_hyperplanes_combined} illustrates the corresponding separating hyperplanes in the feature space for totally positive (TP) and non-totally positive (non-TP) matrices for all matrix sizes considered thus far.

Despite its interpretability, the linear SVM achieved relatively low classification accuracy when compared to previous results, indicating that a simple linear decision boundary is insufficient to separate TP matrices from non–TP ones in this reduced feature space.

This visual evidence is consistent with the relatively low classification accuracy of the linear SVM, confirming that the relationship between the characteristic polynomial coefficients and total positivity is not linear.

\subsection{Nonlinear Kernel (RBF)}
In contrast, when using the RBF kernel, which allows for nonlinear separation by mapping the data into a higher-dimensional feature space, the classification accuracy increased substantially. The informative content of this comparison is not the bare fact that the relationship is nonlinear (which is immediate, since total positivity is defined by a system of polynomial inequalities), but rather its specific, low-complexity geometric character. In the reduced three-dimensional coefficient space the two classes are \emph{not} linearly separable: a single hyperplane plateaus at accuracies of $0.877$, $0.837$ and $0.805$ for $n=5,10,30$ (Table~\ref{tab:svm_acc}). Yet a low-degree nonlinearity already suffices: the RBF kernel reaches accuracies near $0.99$, and, as shown in Section~\ref{sec:ellipsoid}, a single quadratic surface, a Mahalanobis ellipsoid, already captures the region occupied by the TP samples. The substantive observation is thus the gap between linear ($\approx 0.85$) and quadratic/RBF ($\approx 0.99$) separability, indicating a decision boundary that is nonlinear but geometrically simple.

\begin{table}[p]
\centering
\caption{Classification accuracy for SVMs with linear and RBF kernels.}
\label{tab:svm_acc}
\begin{tabular}{lcc}
\hline
\textbf{Matrix size} & \textbf{Linear kernel} & \textbf{RBF kernel} \\
\hline
5×5  & 0.877 & 0.9945 \\
10×10 & 0.8371 & 0.9962 \\
30×30 & 0.8051 & 0.9885 \\
\hline
\end{tabular}
\end{table}

\section{Geometric Characterization via Mahalanobis Ellipsoids}
\label{sec:ellipsoid}
As a preliminary step to characterize the distribution, a three-dimensional scatter representation was generated using the three highest coefficients of the characteristic polynomial for each matrix. Observation of the regions occupied by TP and non-TP samples shows that TP matrices tend to concentrate within a localized, well-defined region of the space,  whereas non-TP matrices exhibit a more dispersed distribution. This distinction indicates a degree of separability between the two classes, suggesting that the TP samples occupy a bounded region of the feature space that may be distinguishable through nonlinear decision boundaries.

\subsection{Mahalanobis Distance and Ellipsoids}

To formally describe the geometric region occupied by the totally positive (TP) matrices in the coefficient space, we model their distribution using Mahalanobis ellipsoids. Let
\[
x \in \mathbb{R}^3
\]
be the vector formed by the three highest-order coefficients of the charactheristic polynomial of a matrix. Given a set of such vectors corresponding to TP samples, their mean and covariance matrix
\[
\mu \in \mathbb{R}^3, \qquad
\Sigma \in \mathbb{R}^{3 \times 3}
\]
are calculated.

The Mahalanobis distance of a point x from this distribution is defined as
\[
D_M^2(x) = (x - \mu)^T \Sigma^{-1} (x - \mu).
\]

A Mahalanobis ellipsoid of radius \(\tau > 0\) is then given by the set
\[
E(\mu, \Sigma, \tau)
=
\left\{
    x \in \mathbb{R}^3 :
    (x - \mu)^T \Sigma^{-1} (x - \mu) \le \tau
\right\}.
\]

In this formulation, \(\mu\) specifies the center of the ellipsoid, representing the average location of the TP samples in the coefficient space; \(\Sigma\) determines the shape and orientation of the ellipsoid, representing the correlations between coefficients; and \(\tau\) controls the size of the ellipsoid, determining how much of the TP distribution is enclosed. This definition provides a precise geometric description of the region in which the TP matrices concentrate, forming the basis for the convergence and numerical analyses later in the section.

Unlike the Euclidean distance, the Mahalanobis metric accounts for correlations between features and the overall shape of the data cloud, making it ideal for identifying regions of concentration with different values along different axes. The resulting constant-distance contours form ellipsoids that enclose points of equal probability density under the TP feature distribution.

\begin{figure}[tbp]
    \centering

    \begin{subfigure}{0.8\textwidth}
        \centering
        \IfFileExists{mahalanobis_ellipsoid_5x5.pdf}
        {\includegraphics[width=\textwidth]{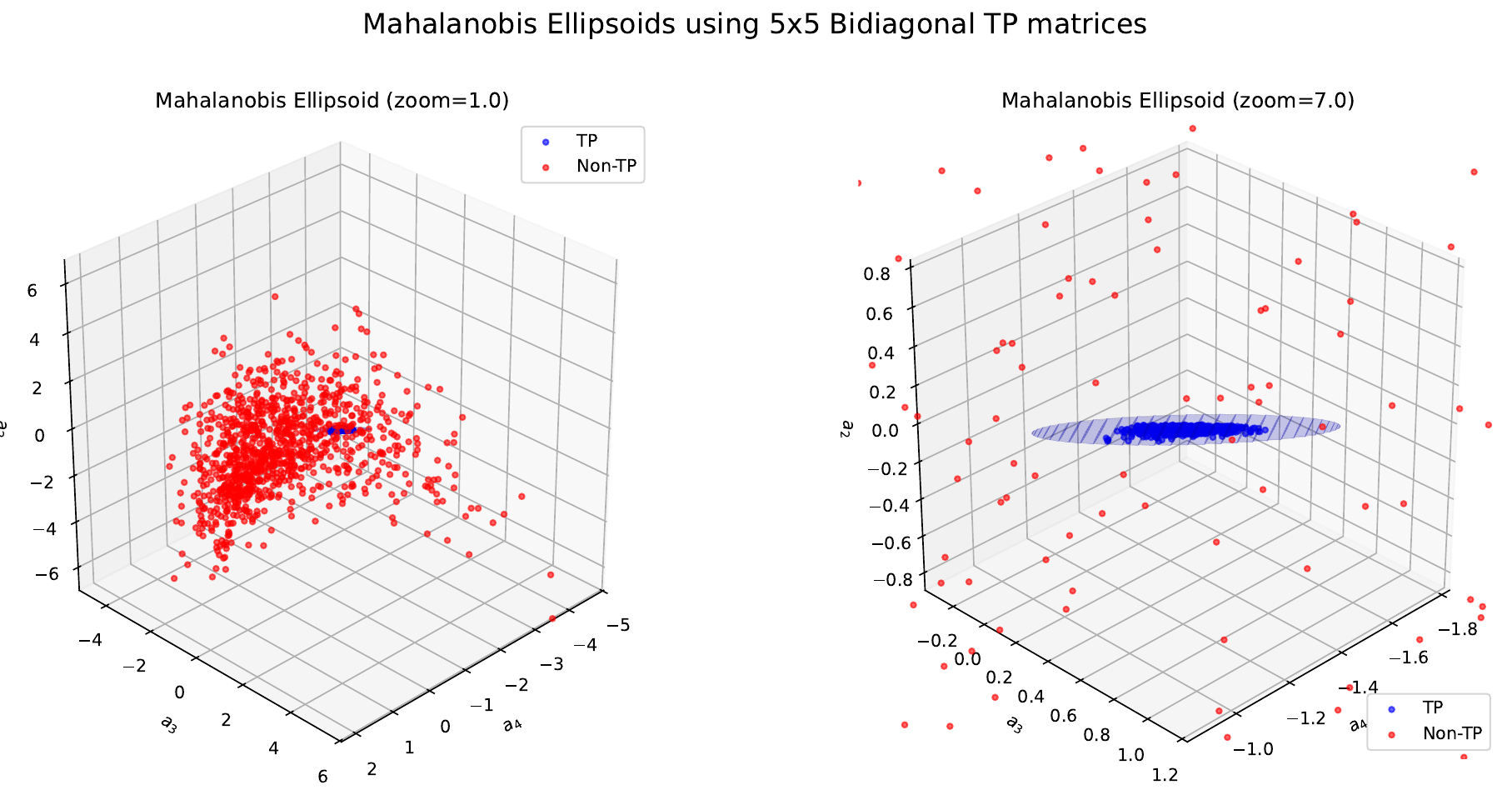}}
        {\fbox{\parbox[c][5cm][c]{\textwidth}{\centering Placeholder for missing image}}}
        \label{fig:mahalanobis_ellipsoid_5x5}
    \end{subfigure}

    \begin{subfigure}{0.8\textwidth}
        \centering
        \IfFileExists{mahalanobis_ellipsoid_10x10.pdf}
        {\includegraphics[width=\textwidth]{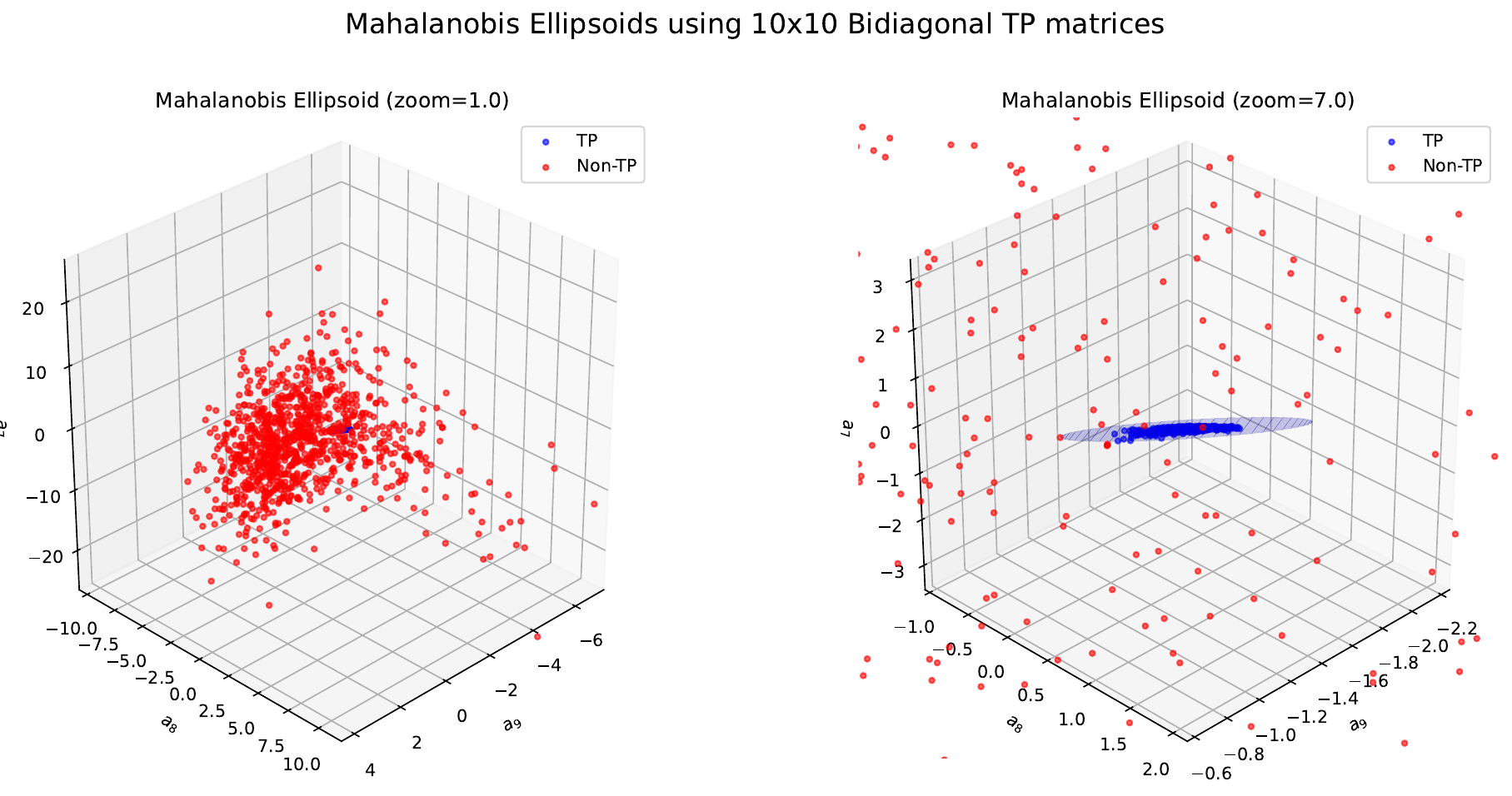}}
        {\fbox{\parbox[c][5cm][c]{\textwidth}{\centering Placeholder for missing image}}}
        \label{fig:mahalanobis_ellipsoid_10x10}
    \end{subfigure}

    \begin{subfigure}{0.8\textwidth}
        \centering
        \IfFileExists{mahalanobis_ellipsoid_30x30.pdf}
        {\includegraphics[width=\textwidth]{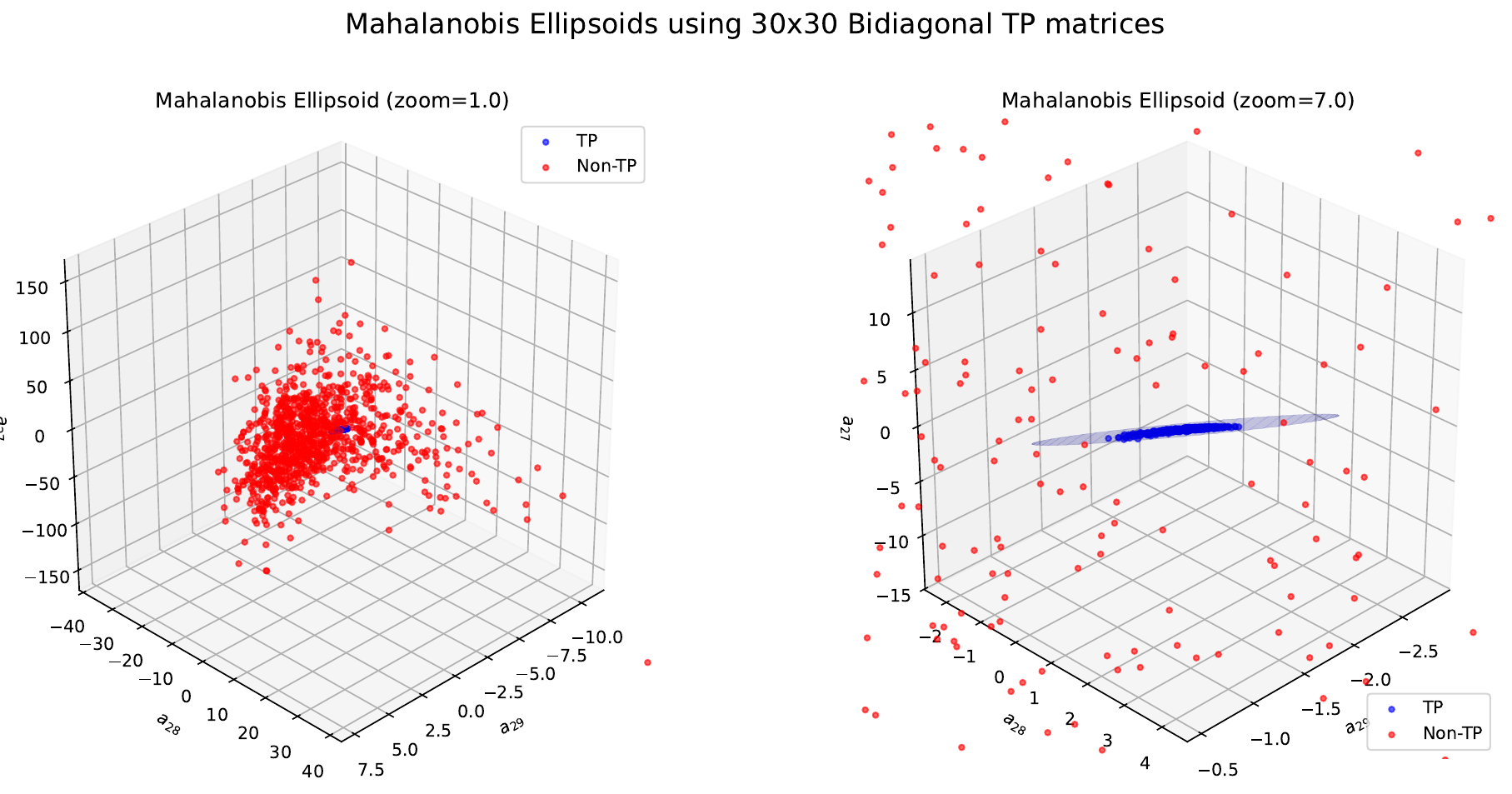}}
        {\fbox{\parbox[c][5cm][c]{\textwidth}{\centering Placeholder for missing image}}}
        \label{fig:mahalanobis_ellipsoid_30x30}
    \end{subfigure}

    \caption{Mahalanobis ellipsoids for TP matrices of sizes $5\times 5$, $10\times 10$ and $30\times 30$ in the space of the three highest characteristic polynomial coefficients.}
    \label{fig:mahalanobis_ellipsoids_all}
\end{figure}

With this model in place, we compute the Mahalanobis distance for each point relative to the distribution of TP matrices. As shown in Figure~\ref{fig:mahalanobis_ellipsoids_all}, the ellipsoids capture all TP matrices, while most non-TP matrices lie outside the boundaries. This confirms that the TP samples occupy a distinct and statistically coherent region, whose boundaries are defined by the correlation structure between the coefficients.

\subsection{Convergence Analysis of the Ellipsoid Parameters}

To verify that the parameters defining the ellipsoid are not a result of the particular dataset used, a convergence study was carried out based on repeated subsampling of the TP matrices. The procedure begins by calculating a reference ellipsoid using the full set of available TP samples (e.g. 50000 samples). The resulting parameters are $\mu_{\mathrm{ref}}$ and $\Sigma_{\mathrm{ref}}$.

To assess the stability of these parameters, several subsample sizes \(N\) were selected from the range \(1000, 2000, 3000, \dots, 49000\), and for each value of \(N\), the TP dataset was randomly resampled multiple times. For every subsample, a new pair of parameters \((\mu_{\mathrm{sub}}, \Sigma_{\mathrm{sub}})\) was computed, yielding a collection of estimated ellipsoids for each fixed sample size.

The convergence toward the reference ellipsoid was quantified by comparing these subsample parameters with the reference ones. The discrepancy in the center was measured using the Euclidean norm,
\[
\|\mu_{\mathrm{sub}} - \mu_{\mathrm{ref}}\|,
\]
while the discrepancy in the covariance matrix was measured using the Frobenius norm,
\[
\|\Sigma_{\mathrm{sub}} - \Sigma_{\mathrm{ref}}\|_F,
\]
For each subsample size, the mean and standard deviation of these errors across the repetitions were recorded

\begin{figure}[tbp]
    \centering

    \begin{subfigure}{0.8\textwidth}
        \centering
        \IfFileExists{convergence_5x5.pdf}
            {\includegraphics[width=\textwidth]{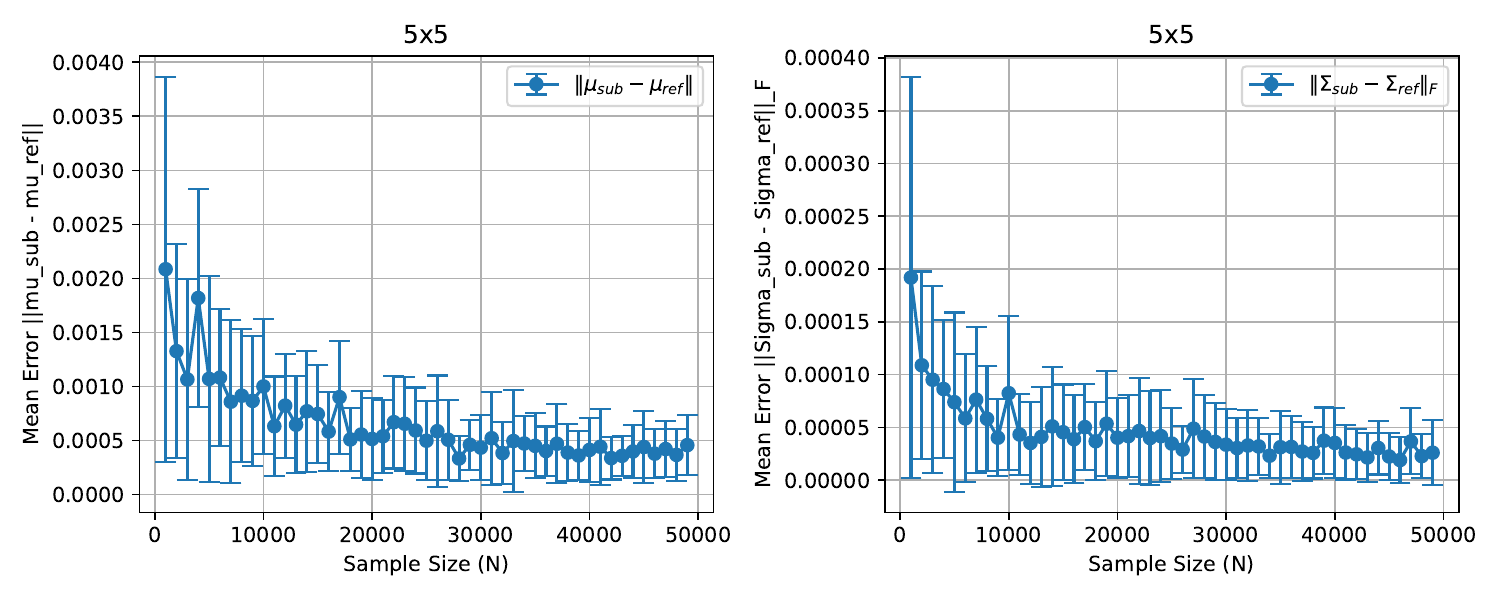}}
            {\fbox{\parbox[c][4cm][c]{\textwidth}{\centering Missing 5x5 convergence plot}}}
        \label{fig:conv_5x5}
    \end{subfigure}

    \begin{subfigure}{0.8\textwidth}
        \centering
        \IfFileExists{convergence_10x10.pdf}
            {\includegraphics[width=\textwidth]{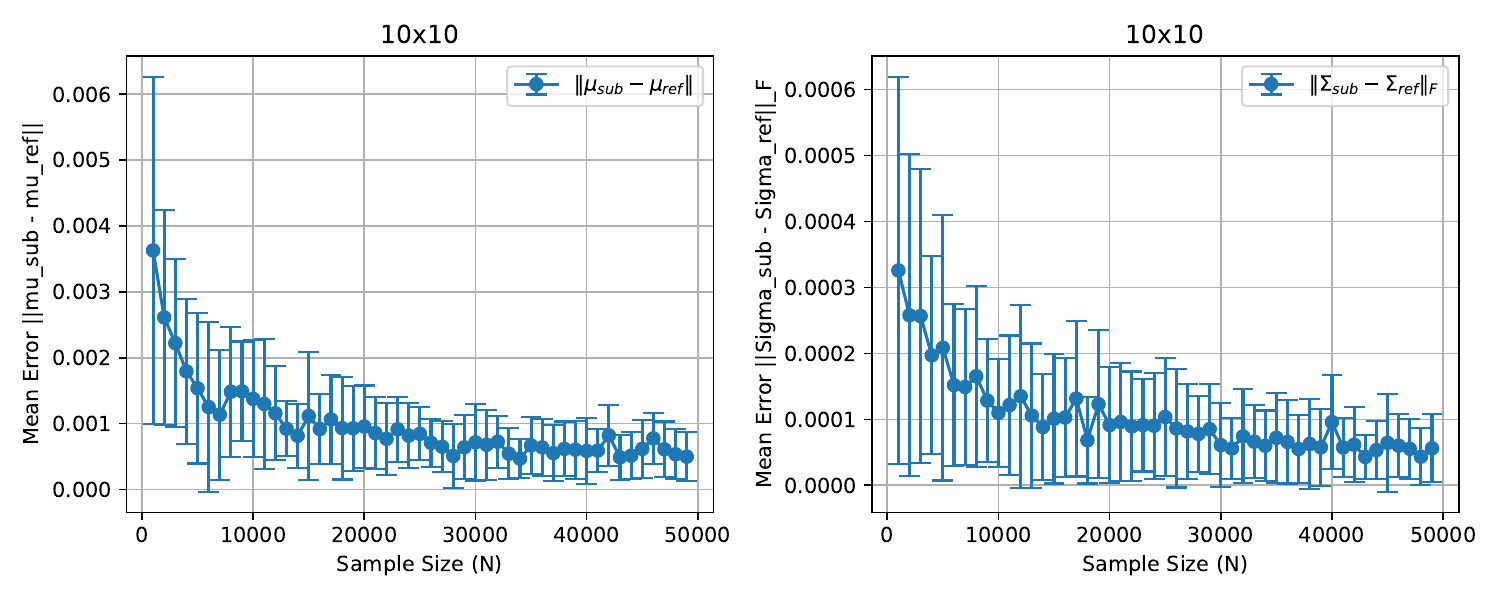}}
            {\fbox{\parbox[c][4cm][c]{\textwidth}{\centering Missing 10x10 convergence plot}}}
        \label{fig:conv_10x10}
    \end{subfigure}

    \begin{subfigure}{0.8\textwidth}
        \centering
        \IfFileExists{convergence_30x30.pdf}
            {\includegraphics[width=\textwidth]{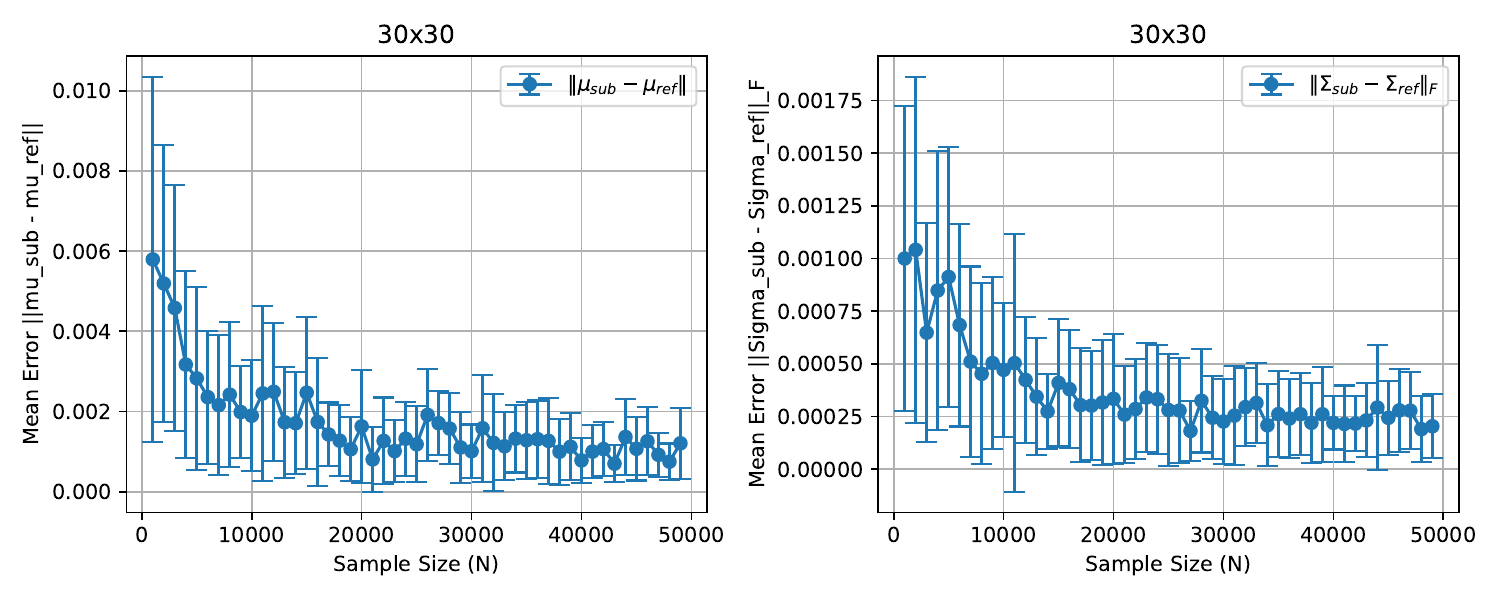}}
            {\fbox{\parbox[c][4cm][c]{\textwidth}{\centering Missing 30x30 convergence plot}}}
        \label{fig:conv_30x30}
    \end{subfigure}

    \caption{Convergence of the Mahalanobis ellipsoid parameters across different matrix dimensions. Each plot shows the decay of mean and covariance errors as the subsample size $N$ increases, demonstrating the stability of the fitted ellipsoid.}
    \label{fig:conv_all}
\end{figure}

The resulting plots, shown in Figure~\ref{fig:conv_all}, display the dependence of both the mean error and the covariance error on the subsample size $N$ for the $5\times 5$, $10\times 10$, and $30\times 30$ cases, respectively. In all dimensions, the same trend is observed: as the subsample increases, the estimates $(
\mu_{\mathrm{sub}}, \Sigma_{\mathrm{sub}})$ progressively approach the reference parameters $(
\mu_{\mathrm{ref}}, \Sigma_{\mathrm{ref}})$. The consistent behaviour across all three matrix dimensions indicates that the fitted ellipsoid is statistically robust.

\subsection{Other Totally Positive Families}

To assess whether the observed patterns in the coefficient space are intrinsic to total positivity rather than specific to a single matrix construction, we extended our analysis beyond the bidiagonal family and considered the two additional TP classes, the Cauchy and Vandermonde families, described in section~\ref{sec:dataset}.

For each matrix family, we sampled a collection of matrices of identical size and computed the coefficients of their characteristic polynomials. The Mahalanobis ellipsoids associated with each family was constructed from the corresponding mean vector and covariance matrix. Table~\ref{tab:ellipsoid_params} reports the estimated values of
\[
(\mu_{a_{29}},\, \mu_{a_{28}},\, \mu_{a_{27}}) \quad \text{and} \quad \tau
\]
for the bidiagonal, Vandermonde and Cauchy families in dimension $30 \times 30$, providing a quantitative summary of the geometric location and scale of each distribution in the coefficient space.

\begin{table}[p]
    \centering
    \caption{Estimated parameters $(\mu_{a_{29}}, \mu_{a_{28}}, \mu_{a_{27}})$ and $\tau$ for the three TP matrix families in dimension $30\times 30$.}
    \label{tab:ellipsoid_params}
    \begin{tabular}{lcccc}
    \toprule
    Family & $\mu_{a_{29}}$ & $\mu_{a_{28}}$ & $\mu_{a_{27}}$ & $\tau$ \\
    \midrule
    Bidiagonal & -1.6822083 & 1.0165671 & -0.2853082 & 122.426 \\
    Vandermonde & -2.3714268e-01 & 1.4417382e-02 & -1.4372400e-04 & 251.705 \\
    Cauchy & -1.1928841 & 0.23345031 & -0.00724995 & 138.201 \\
    \bottomrule
    \end{tabular}
\end{table}

\begin{figure}[tbp]
    \centering

    \begin{subfigure}{0.45\textwidth}
        \centering
        \IfFileExists{tp_ellipsoids_5x5.pdf}
        {\includegraphics[width=\textwidth]{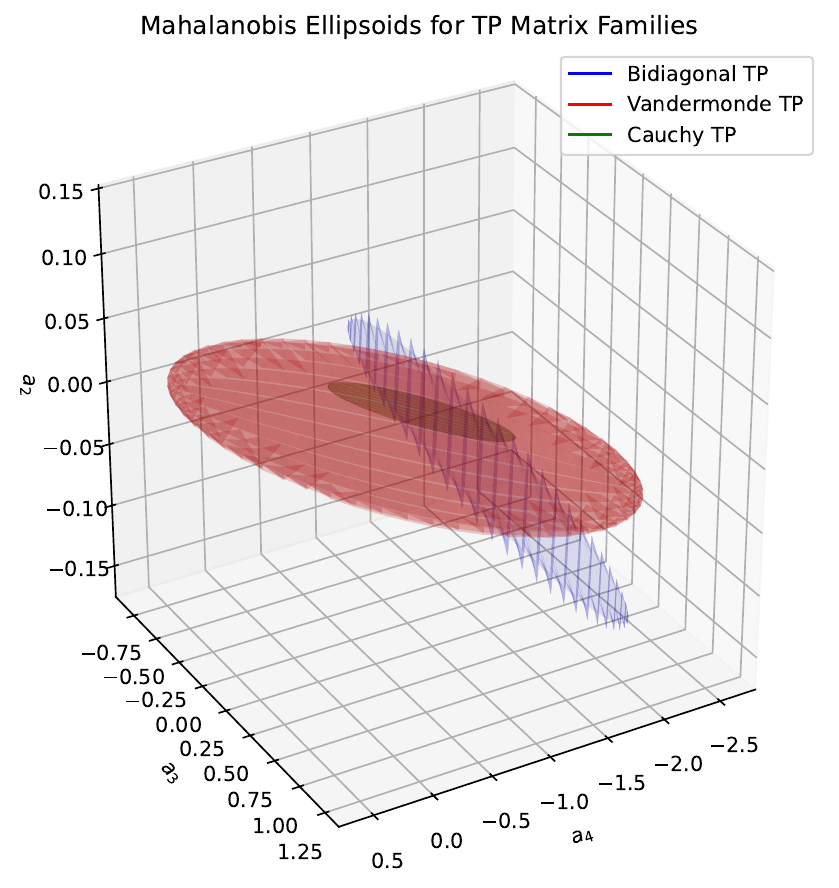}}
        {\fbox{\parbox[c][4cm][c]{\textwidth}{\centering Placeholder for 5x5 image}}}
        \subcaption{Mahalanobis ellipsoids for the $5\times 5$ case.}
        \label{fig:tp_ellipsoids_5x5}
    \end{subfigure}
    \hfill
    \begin{subfigure}{0.45\textwidth}
        \centering
        \IfFileExists{tp_ellipsoids_10x10.pdf}
        {\includegraphics[width=\textwidth]{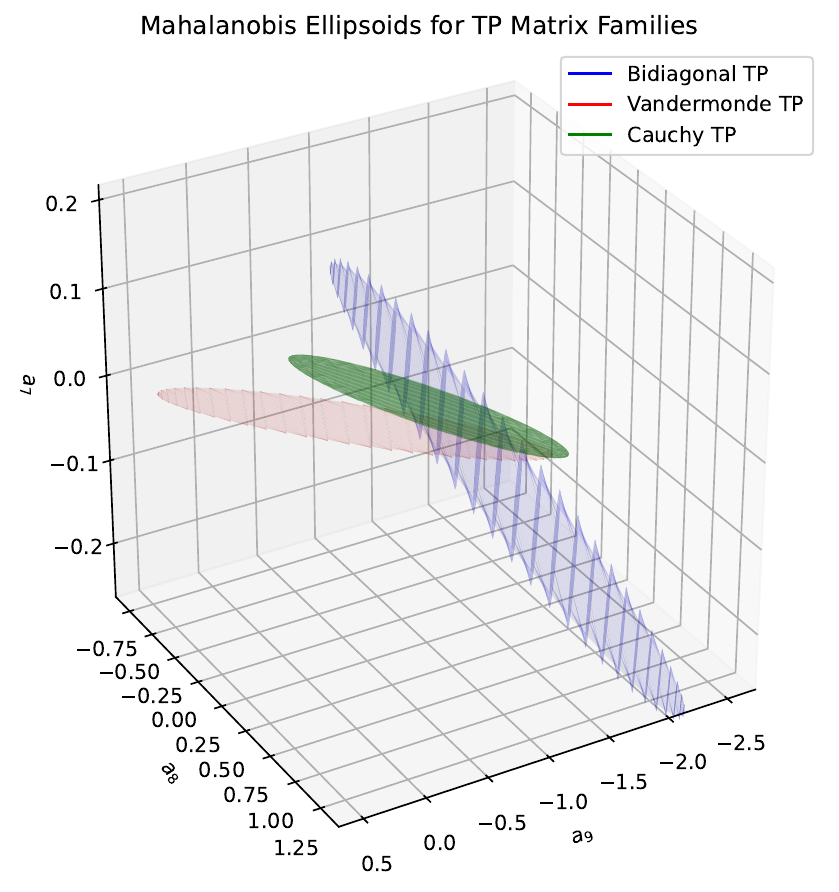}}
        {\fbox{\parbox[c][4cm][c]{\textwidth}{\centering Placeholder for 10x10 image}}}
        \subcaption{Mahalanobis ellipsoids for the $10\times 10$ case.}
        \label{fig:tp_ellipsoids_10x10}
    \end{subfigure}

    \begin{subfigure}{0.45\textwidth}
        \centering
        \IfFileExists{tp_ellipsoids_30x30.pdf}
        {\includegraphics[width=\textwidth]{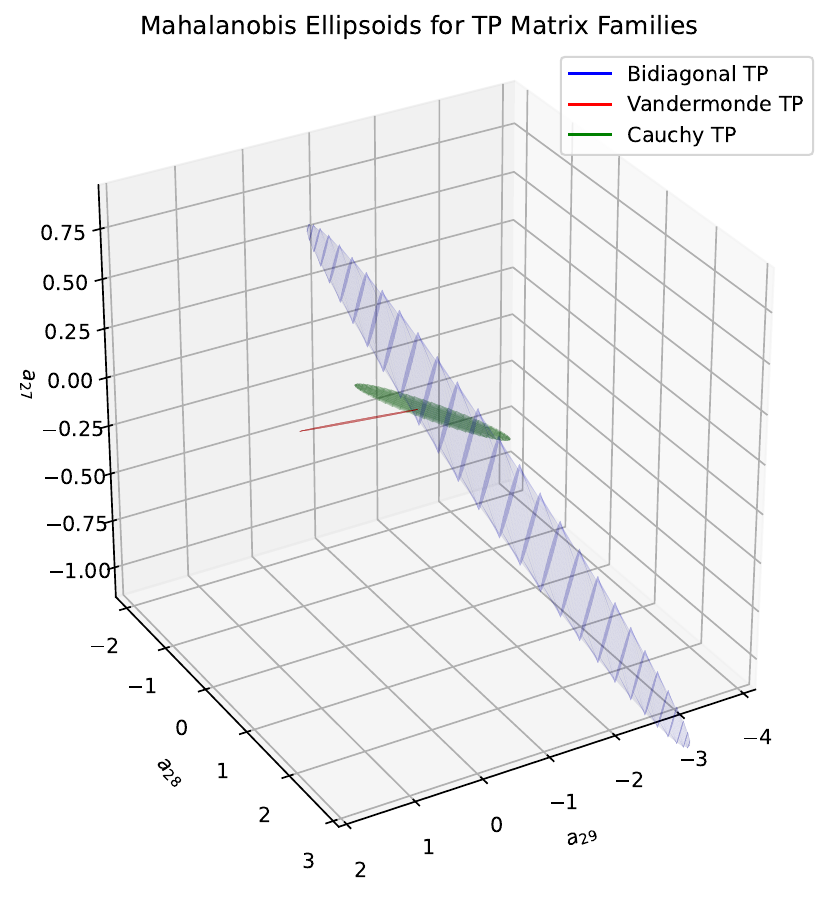}}
        {\fbox{\parbox[c][4cm][c]{\textwidth}{\centering Placeholder for 30x30 image}}}
        \subcaption{Mahalanobis ellipsoids for the $30\times 30$ case.}
        \label{fig:tp_ellipsoids_30x30}
    \end{subfigure}

    \caption{Mahalanobis ellipsoids for TP matrices from the Bidiagonal, Vandermonde, and Cauchy families for matrix sizes $5\times 5$, $10\times 10$ and $30\times 30$.}
    \label{fig:tp_ellipsoids_all}
\end{figure}

Analysis of the results, shown in Figures~\ref{fig:tp_ellipsoids_5x5}-\ref{fig:tp_ellipsoids_30x30}, revealed a crucial geometric trend: each TP matrix family generated a distinct ellipsoidal signature, confirming that the coefficient representation is consistent across different algebraic constructions.

\begin{figure}[tbp]
\centering
\includegraphics[width=\textwidth]{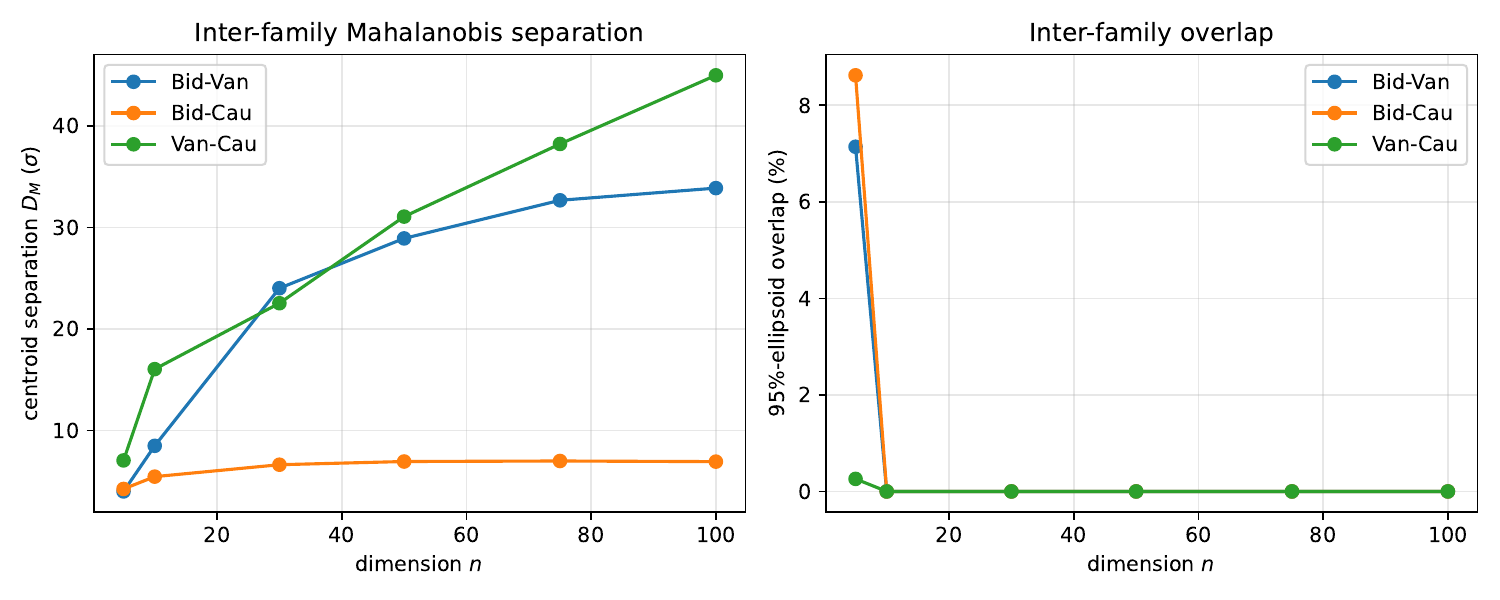}
\caption{Inter-family separation as a function of the dimension $n$. Left: Mahalanobis separation $D_M$ of the family centroids (in standard deviations). Right: worst-case overlap, the percentage of one family lying inside another family's $95\%$ Mahalanobis ellipsoid. The overlap vanishes by $n=10$ while $D_M$ grows monotonically.}
\label{fig:separation}
\end{figure}

Significantly, the degree of inter-family separation increased dramatically with the matrix dimension. For $n=5$ (Figure~\ref{fig:tp_ellipsoids_5x5}), the ellipsoids exhibit a notable degree of overlap. However, as the dimension grew to $10$ and $30$ (Figures~\ref{fig:tp_ellipsoids_10x10} and \ref{fig:tp_ellipsoids_30x30}), the unique signatures became progressively more distinct in location, shape, and orientation. For the $30 \times 30$ matrices, the ellipsoids are effectively disjoint, as confirmed by a numerical search finding a negligible (or zero) intersection region. This verification was performed by constructing a bounding box around the union of the distributions and sampling a dense grid of points ($50^3$ samples) to test for simultaneous inclusion in multiple Mahalanobis ellipsoids.
To make this dimensional dependence quantitative, we measured the separation between families by two scale-aware quantities. For each pair of families we computed the Mahalanobis separation of their centroids with respect to the pooled covariance,
\[
D_M(\mathcal{F},\mathcal{G})=\sqrt{(\mu_{\mathcal F}-\mu_{\mathcal G})^{\!\top}\,\Sigma_{\mathrm{pool}}^{-1}\,(\mu_{\mathcal F}-\mu_{\mathcal G})},\qquad \Sigma_{\mathrm{pool}}=\tfrac12(\Sigma_{\mathcal F}+\Sigma_{\mathcal G}),
\]
which expresses the centroid distance in standard-deviation units, and the \emph{overlap}, defined as the worst-case fraction of the points of one family that fall inside the other family's $95\%$ ellipsoid ($\tau=\chi^2_{3,0.95}$; the choice of level is not essential, since other high levels such as $90\%$ or $99\%$ give the same picture). Table~\ref{tab:separation} reports both quantities for $n\in\{5,10,30,50,75,100\}$, and Figure~\ref{fig:separation} displays them. The pairwise overlap is already $0\%$ at $n=10$ and remains $0\%$ through $n=100$, so the families are effectively disjoint already in moderate dimension. The centroid separations $D_M$ stay large throughout: they grow to several tens of standard deviations for the Bidiagonal--Vandermonde and Vandermonde--Cauchy pairs and stabilize around $7$ standard deviations for the Bidiagonal--Cauchy pair, never approaching the overlap regime.

\begin{table}[tbp]
\centering
\caption{Inter-family separation versus dimension. $D_M$ is the Mahalanobis separation of the centroids (in standard deviations); the overlap is the worst-case percentage of one family inside the other's $95\%$ ellipsoid. Bi, Va, Ca denote the Bidiagonal, Vandermonde and Cauchy families. Each entry is estimated from $5000$ random matrices per family; the values are sampling estimates, but the qualitative pattern (vanishing overlap beyond $n=10$, and separations growing for two pairs while stabilizing for the third) is stable across resampling.}
\label{tab:separation}
\begin{tabular}{c ccc ccc}
\toprule
 & \multicolumn{3}{c}{$D_M$ (std.\ dev.)} & \multicolumn{3}{c}{overlap (\%)}\\
\cmidrule(lr){2-4}\cmidrule(lr){5-7}
$n$ & Bi--Va & Bi--Ca & Va--Ca & Bi--Va & Bi--Ca & Va--Ca\\
\midrule
$5$   & $4.0$  & $4.3$  & $7.1$  & $7.1$ & $8.6$ & $0.3$\\
$10$  & $8.5$  & $5.5$  & $16.1$ & $0$ & $0$ & $0$\\
$30$  & $24.0$ & $6.6$  & $22.6$ & $0$ & $0$ & $0$\\
$50$  & $28.9$ & $7.0$  & $31.1$ & $0$ & $0$ & $0$\\
$75$  & $32.7$ & $7.0$  & $38.2$ & $0$ & $0$ & $0$\\
$100$ & $33.9$ & $6.9$  & $45.0$ & $0$ & $0$ & $0$\\
\bottomrule
\end{tabular}
\end{table}

The robustness of this geometric separation, persisting and intensifying as the dimension increases, suggests that it is not an artifact of sampling but rather an intrinsic structural property of the coefficient space. There is, moreover, a natural route towards a rigorous statement. By the identity $a_{n-k}=(-1)^kE_k$ recalled in the Introduction, the projected cloud of a family is governed by the joint distribution of $(E_1,E_2,E_3)$ of its (Frobenius-normalized) random model, and for each family these are explicit symmetric functions of the generating parameters (the bidiagonal multipliers, the Vandermonde nodes, or the Cauchy points). A proof of this disjointness would then consist of two ingredients: (i) computing the limiting centroids $\mu^{(\cdot)}_n=\mathbb{E}(E_1,E_2,E_3)$ and covariances of each family and showing that the resulting clouds remain separated; and (ii) a concentration argument bounding the fluctuations of $(E_1,E_2,E_3)$ around their means, so that the projected clouds remain confined to shrinking neighbourhoods of their centres. We regard establishing this as the main open problem suggested by our experiments. Motivated by the quantitative evidence above, we formulate the following conjecture.
\begin{conjecture}[Ellipsoidal separation of TP families]
Let $\mathcal{B}_n, \mathcal{V}_n, \mathcal{C}_n \subset \mathbb{R}^{n \times n}$ denote the Bidiagonal-product, Vandermonde, and Cauchy TP matrix families generated by the random models of Section~\ref{sec:dataset}, and let $\pi: \mathbb{R}^{n \times n} \to \mathbb{R}^3$ map a matrix $A$ to the three highest-order coefficients $(a_{n-1}, a_{n-2}, a_{n-3})$ of its characteristic polynomial. For a level $\alpha\in(0,1)$ and a family, let $E^{\alpha}(n)\subset\mathbb{R}^3$ be the Mahalanobis ellipsoid of the corresponding projected cloud that encloses a fraction $\alpha$ of its samples. Then, for every fixed $\alpha\in(0,1)$, there exists a dimension $n_0(\alpha)$ such that, for all $n \ge n_0(\alpha)$, the three ellipsoids $E^{\alpha}_{\mathcal{B}}(n), E^{\alpha}_{\mathcal{V}}(n), E^{\alpha}_{\mathcal{C}}(n)$ are pairwise disjoint, and their pairwise separation, measured in units of the cloud spread, stays bounded below by a positive constant as $n$ grows. In other words, each family occupies its own convex, approximately ellipsoidal region of the coefficient space, and these regions do not merge as the dimension increases.
\end{conjecture}

Overall, the cross-family validation demonstrates that, despite their distinct algebraic definitions, Cauchy, Vandermonde, and bidiagonal matrices exhibit consistent geometric patterns in the coefficient space. The observation that these patterns manifest as unique and separable ellipsoidal signatures reinforces the view that total positivity is a robust and scalable property of the characteristic polynomial coefficients representation, providing a deeper understanding of how algebraic relationships translate into geometric patterns in higher dimensions.

\section{Concluding Remarks}
\label{sec:conclusion}

The analysis conducted in this work leads to several conclusions regarding the structure and discrimination of totally positive (TP) matrices. The findings reveal a compelling hierarchy of separation in the coefficient space. First, the highest-order characteristic polynomial coefficients were found to carry sufficient and concentrated discriminatory information to distinguish TP from non-TP matrices with high accuracy. Second, the boundary for this separation is inherently nonlinear, as confirmed by the RBF-kernel SVM results. Third, distinct TP matrix families, including bidiagonal, Vandermonde, and Cauchy constructions, form unique ellipsoidal clusters within the coefficient space. Finally, this geometric distinction between families becomes significantly stronger and effectively disjoint as the matrix dimension increases, highlighting a spectral signature for each matrix construction.

Collectively, these results show that a small set of interpretable spectral quantities can encode surprisingly rich structural information. Beyond validating the specific approach used here, this has implications for the study of other classes of structured matrices and related mathematical objects. In particular, the work suggests that low-dimensional coefficient representations may provide a useful bridge between classical matrix structure and geometric classification.

The present work also illustrates how exploratory learning tools can be used in support of mathematical investigation. Here, such tools served mainly to identify the coefficients that deserve closer attention, while the main conclusions concern the geometry of the resulting coefficient space and the conjectural separation of structured TP families.

Looking at the bigger picture, the geometric insights uncovered here suggest several directions for future study. One possibility is to investigate whether similar patterns are maintained for larger matrices or for additional structured families beyond the ones considered here. Another promising direction is to examine whether the geometric separation observed in the coefficient space can be described in simpler asymptotic terms as the matrix dimension increases. It may also be worthwhile to explore whether other algebraic features exhibit comparable clustering behavior; natural candidates include other symmetric functions of the spectrum, such as the power sums $\operatorname{tr}(A^k)$ or the remaining lower-order characteristic coefficients, the spectra or traces of the compound matrices $C_k(A)$ (whose entries are the $k\times k$ minors of $A$), which are the natural carriers of minor information since $E_k=\operatorname{tr} C_k(A)$, the multipliers of the bidiagonal (Neville) factorization, and singular-value-based invariants. More generally, extending the datasets and refining the geometric analysis may help clarify how universal these patterns are and whether they reflect deeper principles governing total positivity.

\section*{Declaration of interest}
The authors declare no conflicts of interest.

\section*{Declaration of generative AI and AI-assisted technologies in the manuscript preparation process}
During the preparation of this work, the authors used ChatGPT in order to improve language and organization of the manuscript. After using this tool, the authors reviewed and edited the content as needed and take full responsibility for the content of the published article.

\section*{Data availability statement}
The datasets analyzed in this study consist of generated matrices and derived feature sets produced by the computational procedures described in the manuscript. The code used to generate the structured matrices, to verify total positivity in exact arithmetic, and to reproduce the geometric analysis is available from the corresponding author upon reasonable request.

\section*{Acknowledgements}
Tiago Closs acknowledges scholarship support from CNPq (Conselho Nacional de Desenvolvimento Cient\'ifico e Tecnol\'ogico), Brazil.

\bibliographystyle{plain}

\end{document}